\documentclass{article}




\usepackage[final]{neurips_2024}


\usepackage[utf8]{inputenc} 
\usepackage[T1]{fontenc}    
\usepackage{hyperref}       
\usepackage{url}            
\usepackage{booktabs}       
\usepackage{amsfonts}       
\usepackage{nicefrac}       
\usepackage{microtype}      
\usepackage{xcolor}         

\title{Formatting Instructions For NeurIPS 2024}

%

\author{%
  Omer Amichay \\
  Tel Aviv University\\
  \texttt{omeramichay@mail.tau.ac.il} \\
  \And
  Yishay Mansour \\
  Google Research, Tel Aviv University \\
  \texttt{mansour.yishay@gmail.com} \\
}

\maketitle 
\begin{abstract}
We consider non-stationary multi-arm bandit (MAB) where the expected reward of each action follows a linear function of the number of times we executed the action.
Our main result is a tight regret bound of $\tilde{\Theta}(T^{4/5}K^{3/5})$, by providing both upper and lower bounds. 
We extend our results to derive instance dependent regret bounds, which depend on the unknown parametrization of the linear drift of the rewards.
\end{abstract}

\section{Introduction}
The multi-armed bandit (MAB) problem serves as a fundamental framework in decision-making under uncertainty,  spanning various domains in machine learning.
At the heart of the MAB problem lies the delicate balance between exploration and exploitation, where at each step, the agent decides which arm to pull, seeking to maximize cumulative rewards over time. 
There are many real world applications for MAB, such as online advertising, recommendations and more. 
In much of the literature, the rewards of the MAB are assumed to be stochastic and stationary (see, \citet{slivkins2019introduction,lattimore2020bandit}).

In the non-stationary MAB, the payoff associated with each arm may change over time.
An extreme case is an adversarial environment, where the reward sequence is arbitrary. There is a vast literature on this topic, bounding the regret to the best action, in hindsight. The HEDGE \citet{freund1995desicion}, in the full information case, and EXP3 \citet{auer1995gambling}, in the bandit case , are two classical algorithms for the adversarial setting.

There are also non-stationary stochastic setting, such as, persistent drift \citet{freund1997learning}, Brownian motion \citet{slivkins2008adapting}, uncertainty over timing of rewards distributions changes \citet{garivier2008upper}, bounded reward variation \citet{besbes2014stochastic}, and more. A general class of non-stationary stochastic MAB, where the payoffs changes are coordinated or correlate are the rested and restless bandits. In the rested bandit, the payoff of an action depends on the number of time it was executed, and in the restless bandit the reward depends on the time step (see, \citet{gittins2011multi}).

The Rested MAB (RMAB) framework captures the phenomenon of monotonically changing efficiency with continued execution of the action. For instance, increased experience in performing a given action may lead to improved effectiveness over time. 
Under RMAB we have two different settings, depending whether the rewards decrease of increase. 
The Rotting RMAB problem abstracts 
rewards which decrease monotonically with the number of pulls (\citet{levine2017rotting,seznec2019rotting}).
The Rising RMAB \citet{heidari2016tight}, captures 
rewards that increases monotonically. 

Rising rested MAB with linear drift is the case that the non-stationary expected rewards of each arm have a linear parameterization.
To see the challenges in the model, consider two identical arms with linear increasing rewards. If we play each arm about half the times, we might get a linear regret. In contrast, for a stationary MAB with two identical arms,  any policy have zero regret.
As a motivation we can consider hyperparameter search over multiple algorithms. As we are modifying parameters of each algorithm, we expect the overall system performance to improve. By selecting between the algorithms we are exploring to find the best overall algorithm.
Having a linear rate of increase is a natural modeling assumption, both as a starting step for the theory and a simplistic empirical model.
    




It is customary to measure the performance of online algorithms using the notion of regret, the performance difference between the online algorithm's cumulative rewards and that of a benchmark.
We highlight three common notions of regret's benchmarks, \emph{static}, \emph{dynamic} and \emph{policy} regret.
Static regret uses the performance of the best fixed arm as a benchmark. Dynamic regret uses as a benchmark the performance of the  best sequence of arms.  Policy regret \citep{arora2012online} uses the best policy and allows the environment to react to the policy.

\subsection{Our results}
In this paper, we study the regret of Rising Rested MAB with linear drift, i.e., the expectation of the reward function of each arm is linear with respect to the number of times the arm was played.
Our main results:
\begin{itemize}
\item We show that the dynamic and static regrets are identical for rising rested MAB with linear drift.
    \item We design R-ed-EE (Rested Explore Exploit) algorithm,
    and show a regret bound of
    $O(T^\frac{4}{5}(\Phi K)^\frac{3}{5}\ln(\Phi KT)^\frac{1}{5})$, where $T$ is the number of time steps, $K$ is the number of actions and $\Phi$ bounds the expectation of the arm rewards.
    \item We design R-ed-AE (Rested Arm Elimination) and HR-re-AE (Halted Rested Arm Elimination) algorithms.
    Algorithm R-ed-AE gives an instance dependent regret bound.
    Algorithm HR-re-AE, extends R-ed-AE, and has, in addition to the instance dependent regret, a worse case $O\b{ T^\frac{4}{5}(\Phi K)^\frac{3}{5}  \ln\b{\Phi K T^2}^{\frac{1}{5}}}$ regret.
    
    \item  
    We show a lower bound of $\Omega(K^\frac{3}{5}T^\frac{4}{5})$ for the regret. This, together with R-ed-EE and HR-re-AE shows tight regret of $\tilde{\Theta}(K^\frac{3}{5}T^\frac{4}{5})$ for the  Rising Rested MAB with linear drift problem. (This is in contrast to the classical $\Theta(\sqrt{T})$ regret bounds in stochastic stationary settings.)
\end{itemize}

\subsection{Related Works}
\paragraph{Stochastic stationary MAB} There is a vast literature on this topic (see, e.g., \citet{slivkins2019introduction}).

\paragraph{Non-Stationary MAB} the Non-Stationary Markovian MAB was studied by \citet{gittins1974dynamic}, who showed that an index policy characterizes the optimal policy for discounted return.
%
 Following  \citet{gittins1974dynamic}, numerous works have explored the domain of rested MAB
(e.g.\citet{whittle1981arm,bertsimas2000restless,nino2001restless}).

\citet{whittle1988restless} introduced the Restless MAB, which also has an optimal index policy.
%
There are works that assume a certain structure to the change in rewards for Restless MAB. 
\citet{freund1997learning} consider persistent distributions drift which is a linear drift.
\citet{slivkins2008adapting} studies a Brownian motion of the rewards. \citet{besbes2014stochastic} have a regret bounded as a function of  the total variation of the rewards. \citet{jia2023smooth} consider smooth non-stationary bandits.

\citet{tekin2012online} introduced rested MAB, where the arm's reward distribution depends on the number of times it was executed.
A special case of rested MAB is \emph{rotting MAB}, where the expected reward of an arm decreases with the number of pulls of the arm (\citet{levine2017rotting, seznec2019rotting}).
\citet{seznec2020single} consider both rested and restless rotting MAB.


\paragraph{Rising rested MAB}
Most related to our problem is the rising rested MAB problem, where the expected rewards of the arms are monotonically non-decreasing in the number of times the decision maker played the arm.
The non-stochastic version of this problem addressed by \citet{heidari2016tight,li2020efficient} for the static regret.
The stochastic  case assumes that the payoffs follows a known parametric form.
The Best Arm Identification framework of this setting was done in 
\citet{cella2021best,mussi2023best}.
%
%
It should be noted that both papers measure the performance of their best arm identification algorithms as a function of the optimal arm and the arm the algorithm chose. In contrast, we measure the dynamic regret with respect to the entire run of the online algorithm when compared with the optimal sequence of arm selection.

\citet{metelli2022stochastic} considered the stochastic rising MAB both for restless and rested, where the payoffs are increasing and concave. For the rested instance the dynamic regret is equivalent to policy regret. 
They designed UCB based algorithms with instance dependent expected regret.
However their algorithm for rested MAB suffers, in some cases, a linear regret. Specifically, for our lower bound instance their algorithm has a linear regret.

\section{Problem Setting and Preliminaries}

In this section we formalize our model of \emph{Rising Rested MAB with Linear Drift.}

\paragraph{Non-stationary rested $K$-MAB}
An instance of a \emph{non-stationary $K$-MAB} is a tuple $(K,T,D)$, where $\K=\{1,2, \ldots ,K\}$ is the set of $K$ arms, $T$ is the time horizon, and for each arm $i$ and $n\leq T$ we have a reward distribution $D_i(n)$ which is the stochastic reward of arm $i$ when it is performed for the $n$-th time, and its expectation is $\mu_i(n)$. We assume that the distributions $D_i(n)$ are $1$-sub-Gaussian, i.e., $\E_{R\sim D_i(n)}[e^{\lambda(R-\mu_i(n))}]\leq e^{\frac{ \lambda^2}{2}}$, for every $\lambda \in \R$.
 Let $\Phi \geq \max_{i\in\K}( \mu_i(T))$ be an upper bound on the maximum expected reward of any arm. (Recall that the expected rewards are non-decreasing.)

At each time $t$ the learner selects an arm $I_t=i$ and observes a reward $R_t\sim D_{i}(N_i(t))$, where $N_{i}(t) = \sum_{\tau=1}^t \mathbbm{1}(I_\tau=i)$ is the number of times arm $i\in \K$ was pulled up to round $t$.

\paragraph{Rested MAB with linear drift}
A Non-stationary rested $K$-MAB has linear drift, if for each arm $i\in \K$ there are 
constants $L_i$ and $b_i$ such that $\mu_i(n)= L_i n + b_i$ 
for every $n\in \{1,2,\ldots,T\}$. 
An instance is called \emph{Rising Rested MAB with linear drift} if $L_i\geq 0$ for every $i\in \K$.

\paragraph{Policies and regret}
A policy $\pi$ selects for each time $t$ a distribution over the arms, i.e.,  $\pi(t)\in\Delta(\K)=\{q\in[0,1]^K | \:\: \sum_i q_i=1\}$. 

The \textbf{static} regret of policy $\pi$ is, 
\[\Re_{stat}=static-Regret(\pi)= \underset{i\in\K}{\max}\sum_{t=1}^{T}\mu_i(t) - \E[\sum_{t=1}^{T}\sum_{i\in \K}\mu_{i}(N_i(t))\mathbb{I}[\pi(t)=i]],\]
where $\mathbb{I}[\cdot]$ is the indicator function.
%
The \textbf{dynamic} regret of policy $\pi$  is,
\[
\Re_{dyn}=
Dynamic-Regret(\pi)= \E[\sum_{t=1}^{T}\mu_{\pi^*(t)}(N_i(t))] - \E[\sum_{t=1}^{T}\sum_{i\in \K}\mu_{i}(N_i(t))\mathbb{I}[\pi(t)=i]],\]
where  $
\pi^* \in \argmax_\pi  \E[\sum_{t=1}^{T}\sum_{i\in \K}\mu_{i}(N_i(t))\mathbb{I}[\pi(t)=i]]
$ is the optimal policy.

\begin{remark}
In our setting the rewards are history dependent, so our dynamic regret is in fact a policy regret. 
\end{remark}

\paragraph{Characterization of the optimal policy}
Theorem \ref{theorem: Rise-Res-LinD-Opt-Policy}, in the supplementary material, shows that the optimal policy for Rising Rested MAB with Linear Drift selects a single arm and always plays it. 
Namely, for $i^* = {\argmax}_{i\in \K}\sum_{t=1}^{T}\mu_i(t)$, \Cref{theorem: Rise-Res-LinD-Opt-Policy} shows that the optimal policy always plays arm $i^*$. 
Therefore, in our setting the dynamic regret is equal to the static regret.

\begin{corollary}\label{cor:opt-arm}
    For Rising Rested MAB with Linear Drift 
    the dynamic regret is equal to the static regret. Namely, the optimal policy plays always arm $i^*.$
\end{corollary}

\paragraph{Notations}
Let $[k,m]=\{k, \ldots, m\} $ and $[m]=\{1, \ldots, m\}$.
Assume we pulled arm $i\in\K$ for $m\in [T]$ times and observed rewards $r_i(1), \ldots , r_i(m)$.
\begin{itemize}

\item The estimated reward, given a window of $[n',n'+M]\subseteq [1,m]$, where $M$ is an even integer, is $\hat{\mu}^M_i (n' +\frac{M}{2})=(1/M)\sum_{n=n'}^{n' + M}r_i(n)$. Note that $\mu_i(n' + \frac{M}{2})=\E[\hat \mu^M_i (n' +\frac{M}{2})]$.\\
(Note that for $\hat{\mu}^M_i (\ell) $ we implicitly have $n'=\ell-M/2$.)

\item
The empirical slope is $\widehat L_i^{2M}= (1/M)\left({\hat\mu_i^M\b{\frac{3M}{2}} - \hat\mu_i^M\b{\frac{M}{2}} }\right)$. Note that $L_i=\E[\widehat L_i^{2M}]$.

 
\item Our estimate for the future rewards $\mu_i(n)$, given $2M$ samples for arm $i$, is
        $\psi^{2M}_i(n) =  \b{\hat\mu_i^M\b{\frac{3M}{2}} + \hat\mu_i^M\b{\frac{M}{2}}}/2 
        + (n - M) \widehat{L}_i^{2M}$. (Note that we might have $n\leq 2M$).
\item 
The cumulative expectation of arm $i \in \K$ during $[n_1,n_2]$ is $s_i(n_1,n_2) = \sum_{n=n_1}^{n_2}\mu_i(n)$, and its estimation
using $2M$ sample points is $\hat{s}_i^{2M}(n_1,n_2) = \sum_{n=n_1}^{n_2}\psi_i^{2M}(n)$.
            
\end{itemize}

We define the following parameters, which will be useful for our concentration bounds:
    Let the confidence parameters be $\gamma_n^{2M} = \sqrt{\frac{\ln\b{\frac{2}{\delta}}}{2M}} + \left | n-M \right |\frac{\sqrt{2\ln\b{\frac{2}{\delta}}}}{M^{1.5}}$,
    and $\Gamma^{2M}(n_1,n_2) = \sum_{n=n_1}^{n_2}\gamma_n^{2M}$.

\section{Algorithm for Rising Rested MAB with  Linear Drift}\label{section: first alg}
In this section we derive an algorithm for the Rising Rested MAB with Linear Drift problem and show that the regret is  $ \tilde \Theta \b{K^{3/5}T^{4/5}}$. This regret bound is tight, as we show in the lower bound of Section \ref{lower bound section}.

The basic idea behind our algorithm is simple. We use an explore-exploit methodology. 
We have an exploration period, in which we explore each arm for $2M$ times. Given the observed rewards, we estimate the model parameters for each arm, namely the slope and the intercept point for the linear function of the arm. 
The surprising outcome is that the dynamic regret bound we get is tight!

In more details.
We are sampling $2M$ times each arm $i \in \K$. In order to recover the line for the rewards of arm $i$ we estimate two points on this line (and they would define the parameters of the line).
The two points are the rewards after $M/2$ and $3M/2$ uses of arm $i$.
Note that the expected reward for $M/2$ is the average of the expected rewards in $[1,M]$, and the expected reward for $3M/2$ is the average of the expected in $[M+1,2M]$.
We denote our estimates as  $\hat\mu_i^M(M/2)$ and $\hat\mu_i^M(3M/2)$.
Given those two estimates, we estimate the slope $\widehat L_i^{2M}$ by their difference divided by $M$.
Using $\hat\mu_i^M(M/2)$, $\hat\mu_i^M(3M/2)$ and $\widehat L_i^{2M}$ we estimate $\hat s_i^{2M}(2M+1,T-2KM)$ the future cumulative rewards of arm $i$ for the remaining time, i.e., next $(T-2KM)$ time steps.
Given our estimates  $\hat s_i^{2M}(2M+1,T-2KM)$ for each arm $i$, we choose to play the arm with the highest estimated value of $\hat s_i^{2M}(2M+1,T-2KM)$. Essentially we are exploiting of the best arm, given our estimates.
(The algorithm appears in Algorithm~\ref{Alg:1}.)

\begin{algorithm}[t]
\caption{R-ed-EE - Rested Explore Exploit}\label{Alg:1}
\begin{algorithmic}[1]
\State Input: $K$, $T$, $M$
\For{$i\in \K$}
    \State Sample arm $i$ for $2M$ times, and observe $r_i(1), \ldots ,r_i(2M)$
    \State $ \hat\mu_i^M\b{\frac{M}{2}} \gets  \frac{1}{M} \sum_{n=1}^{M}r_i(n)$
    \State $ \hat\mu_i^M\b{\frac{3M}{2}} \gets  \frac{1}{M}  \sum_{n=M+1}^{2M}r_i(n)$
    \State $\widehat{L}_i^{2M} \gets \frac{\hat\mu_i^M\b{\frac{3M}{2}} -  \hat\mu_i^M\b{\frac{M}{2}}}{M}$ 
    \Comment{Estimate of the slope}
    \State $\hat{s}_i^{2M}(2M+1,T-2KM) \gets (T-2KM-2M)\sq*{\frac{\hat\mu_i^M\b{\frac{3M}{2}} + \hat\mu_i^M\b{\frac{M}{2}}}{2} 
        + \frac{T-2KM+1}{2}  \widehat{L}_i^{2M}}$
\Statex  \Comment{ computes the sum $\sum_{t=2M+1}^{T-2KM}\psi^{2M}_i(n)$}
\EndFor
\State $\hat{i}^*\gets {\arg\max }_{i\in \K} \: \hat{s}_i^{2M}(2M+1,T-2KM)$ \Comment{The arm with the best estimate future reward}
\For{$t \in [2KM+1,T]$}
    \State Play arm $\hat{i}^*$ 
\EndFor
\end{algorithmic}
\end{algorithm}

\paragraph{Overview of regret analysis}
We define a good event $G$, which bounds the deviations between our estimation and the true values. This holds for the slopes, and the expected rewards. 


\begin{define}\label{good event G instance dependent}
Let $G$ be the good event that after sampling each arm $2M$ times,
for any arm $i\in \K$ :   

$  \av{\hat\mu_i^{M}\b{\frac{M}{2}} - \mu_i\b{\frac{M}{2}}} \leq \sqrt{ \frac{\ln\b{ \frac{2}{\delta}}}{2M}}$ 
    and $  \av{\hat\mu_i^{M}\b{\frac{3M}{2}} - \mu_i\b{\frac{3M}{2}}} \leq \sqrt{ \frac{\ln\b{ \frac{2}{\delta}}}{2M}}$.
\end{define} 

Next, we show that the good event $G$ holds with high probability. 

\begin{lemma}\label{lemma:G}
    The probability of the good even $G$ is at lest $1-2\delta K$ 
\end{lemma}

We then bound the regret using three different terms: (1) The exploration term, which we bound by $2MK$. (2) The exploitation term, which we bound using the confidence bounds of the good event. (3) The low probability event that $G$ does not hold.

\begin{theorem}
For $M=\frac{T^\frac{4}{5}\ln(4\Phi K T)^\frac{1}{5}}{(\Phi K)^\frac{2}{5}} $,    
Algorithm \ref{Alg:1} guarantees regret $O(T^\frac{4}{5}(\Phi K)^\frac{3}{5}\ln(\Phi KT)^\frac{1}{5} )$ for the Rising Rested MAB with Linear Drift.
\end{theorem}
\begin{proof}
     The regret can be partition to three parts.
The first part is during times $[1,2KM]$ when we explore each arm $2M$ times. The second part is during times $[2KM+1,T]$, assuming the good event $G$ holds. The third part is during times $[2KM+1,T]$, when the good event $G$ does not hold. 
%
  
For the first part, i.e., the regret over the first $2M$ samples of each arm, we bound it by $ 2KM\Phi$. 

For the second part, the regret during exploitation stage, assuming the good event $G$ holds. We can upper bound the probability of the good event by one, i.e., $P(G) \leq 1$. Under the good event $G$, by Lemma~\ref{lemma: cumulative error estimation } and Lemma~\ref{cumulative error bound}, we have that for each arm $i \in \K$, we bound the estimation error by,
\[|\hat{s}_i^{2M}(2M+1,T-2KM) - s_i(2M+1,T-2KM)| 
    \leq \Gamma^{2M}(2M+1,T-2KM)
     \leq {T^2\sqrt{\frac{1}{2M^{3}}\ln\b{\frac{2}{\delta}}}}.\]
Note that the regret of the second part is independent of $\Phi$. This is since our confidence intervals do not depend on the magnitude rewards, i.e., $\Phi$.

The third  part bounds the regret when the event $G$ does not hold, i.e., $\bar{G}$ occurs. By Lemma~\ref{lemma:G} we have $P(\bar{G})\leq 2K\delta$, so the regret of this part is bounded by $P(\bar{G})T\Phi\leq 2KT\Phi\delta$.

To summarize,  we have  $\Re\leq 2MK\Phi + \frac{T^2\sqrt{\ln\b{\frac{2}{\delta}}}}{\sqrt{2}M^{1.5}} + 2KT\Phi\delta$. Setting $\delta = \frac{1}{2TK\Phi}$ and $M= \frac{T^\frac{4}{5}\ln(4KT\Phi)^\frac{1}{5}}{(\Phi K)^\frac{2}{5}} $,
    we get that $\Re\leq O(T^\frac{4}{5}(\Phi K)^\frac{3}{5}\ln(\Phi KT)^\frac{1}{5} )$.
\end{proof}

\section{Instance Dependent Upper bound for Rising Rested MAB with Linear Drift}\label{instance dependent section}

In this section we derive two instance dependent algorithms for  Rising Rested MAB with Linear Drift.
The first algorithm gives an instance dependent bound, but in some cases of the parameters, this regret might be linear in $T$. 
%
For this reason we have a second algorithm, for which  we shows that the regret is almost tight (up to a logarithmic factor) in the worse case.

Our first algorithm, essentially, runs an arm elimination methodology, where it eliminate arms who are, with high probability, sub-optimal arms.
Unlike the R-ed-EE algorithm, which samples all arms equally, the arm elimination methodology adjusts the sampling of each arm based on its performance, thereby preventing the over-sampling of sub-optimal arms.

In more detail. The first algorithm R-ed-AE is implementing an arm elimination approach for instance dependent problems. Initially, it starts with the set of all $K$ arms, denoted by $\K'$. In each round the algorithm samples each arm in $i \in \K'$ four times.
The algorithm maintains a counter $N_i$ keeping track over the number of times arm $i$ was sampled. (Note that all non-eliminated arms $i\in \K'$ have approximately the same counter $N_i$.)
Using the observations of arm $i$, the algorithm estimates the line parameters of arm $i$.
The algorithm does so by estimating two points on the line. 
The two points are the rewards after ${N_i}/{4}$ and after ${3N_i}/{4}$ uses of arm $i$.
We denote the estimations as $\hat{\mu}_i^{{N_i}/{2}}(N_i/4)$ and $\hat{\mu}_i^{{N_i}/{2}}(3N_i/4)$, respectively.
Using both estimations we estimate the slope of arm $i$, i.e., $L_i$, denoted  by  $L_i^{N_i}$.
Using the estimations $\hat{\mu}_i^{{N_i}/{2}}({N_i}/{4})$, $\hat{\mu}_i^{{N_i}/{2}}(3{N_i}/{4})$ and $L_i^{N_i}$ we estimate $\hat{s}_i^{N_i}(1,T)$, the expected reward if we played only arm $i$ for the entire $T$ time steps.
At the end of each round for every pair of arms $i,j \in \K' $ we evaluate the difference between the estimations $\hat{s}_i^{N_i}(1,T)-\hat{s}_j^{N_i}(1,T)$. If the difference is greater than $2 \Gamma^{N_i}(1,T)$, then, arm $j$ is eliminated from $\K'$, where $2 \Gamma^{N_i}(1,T)$ is our confidence interval.
We are guaranteed that, with high probability, any eliminated arm is indeed sub-optimal.   

\paragraph{Overview of regret analysis}

We first define the relevant parameters of the instance dependent analysis. Recall that each arm $i$, has expected reward define by $L_i m + b_i=m(L_i+b_i/m)$, where $m$ is the number of times we played arm $i$.
We now would like to define the distinguishability between different arms.
We view $b_i/m$ as a normalized version of the intercept, and later use $b_i/(T+1)$ where we take $m=T+1$.

\begin{define} for any two arm $i,j\in \K$:\\
(1)  $\Delta_{i,j} = b_i - b_j$ the difference of  the intercept points between the arms $i,j$, and $\widetilde{\Delta}_{i,j}=\Delta_{i,j}/(T+1)$.\\
(2)  $L_{i,j} = L_i - L_j$ the difference of the slopes of the two arms.
\end{define}        

\begin{algorithm}[t]
\caption{R-ed-AE- Rested Arm Elimination }
\begin{algorithmic}[1]
\State Input: $K$, $T$, $\delta$
\State Set $N \gets [0]^K$ and $\K' \gets [K]$
\State For each $i \in \K'$ let $B_i$ an empty array,
\State $\tau \gets 0$
\While {$\tau\leq T$ }
        \For{$j\in \K'$ }
            \For{$l \in [4]$}
                \State  sample arm $j$ and set $B_j(N_j)\gets r_j(N_j)$
                \State $N_j \gets N_j + 1$
            \EndFor               
            \State  $ \hat{\mu}_j^\frac{N_j}{2}(\frac{N_j}{4})\gets\frac{2}{N_j}\sum_{n=1}^{\frac{N_j}{2}}B_j(t)$
            \State  $ \hat{\mu}_j^\frac{N_j}{2}(\frac{3N_j}{4})\gets \frac{2}{N_j} \sum_{\frac{N_j}{2}+1}^{N_j}B_j(t)$

            \State Set 
            $ \widehat{L}_j^{N_j} \gets 2\b{\hat{\mu}_i^\frac{N_j}{2}(\frac{3N_j}{4})- \hat{\mu}_i^\frac{N_j}{2}(\frac{N_j}{4})}   /N_j$ 
            \State Set  $\hat{s}_j^{N_j}(1,T) \gets T\sq*{\b{\hat\mu_i^{\frac{N_j}{2}}\b{\frac{3N_j}{4}} + \hat\mu_i^{\frac{N_j}{2}}\b{\frac{N_j}{4}}}/2 
        + \b{\frac{T+1}{2} - M} \widehat{L}_i^{N_j}}$
\Statex  \Comment{computes the sum $\sum_{t=1}^{T}\psi^{N_j}_i(n)$} 
                            
        \EndFor
        \State Set $\tau \gets \tau + 4|\K'|$    
        \For{$i,j \in \K'$} 
            \If {$\hat{s}_i^{N_i}(1,T)-\hat{s}_j^{N_i}(1,T) > 2 \Gamma^{N_i}(1,T)$ }
                \State $\K' \gets \K' \setminus \{j\}$
            \EndIf
        \EndFor        
\EndWhile
\State return $\K'$
\end{algorithmic}
\end{algorithm}

For the analysis we first define a good event $G$, which will hold with high probability. Intuitively, the good event states that all of our estimates are accurate. Formally, the definition of the good event $G$ is as follows.

\begin{define}\label{good event G instance dependent AE}
Let $G$ be the good event that
for all $i\in \K$ and $m\in [T]$: 
$\av{\hat\mu_i^{\frac{m}{2}}(\frac{m}{4})-\mu_i(\frac{m}{4})}\leq\sqrt{\frac{\ln\b{\frac{2}{\delta}}}{m}}$
, $\av{\hat\mu_i^{\frac{m}{2}}(\frac{3m}{4})-\mu_i(\frac{3m}{4})} \leq\sqrt{\frac{\ln\b{\frac{2}{\delta}}}{m}}$  
and $\av{ \widehat{L}_i^{m}-L_i}\leq \frac{ 2\sqrt{\ln\b{\frac{2}{\delta}}} } {m^{1.5}}$.
\end{define} 

The following claims that the good event holds with high probability.  

\begin{lemma}\label{lemma: event G for AE}
    The probability of the good even $G$ is at lest $1-2TK\delta$.
\end{lemma}

From now on we will assume that the good event $G$  holds.

We start by showing that, under the good event $G$, the best arm $i^*$ is never eliminated. (Note that we fixed the best arm, initially.)
\begin{lemma}\label{Lemma: best ar never eliminated}
    Under $G$ we never eliminate the optimal arm $i^*$.
\end{lemma}

Next we bound the number of times we can play any sub-optimal arm. The sample size of the sub-optimal arm $j$ depend only on the parameters $ \widetilde{\Delta}_{i^*,j}$ and $ L_{i^*,j}$. 

\begin{lemma} \label{elimination Lemma R-ed-AE}
Algorithm R-ed-AE samples a sub-optimal arm $j$ at most
$\sq*{\frac{  16 \sqrt{\ln\b{\frac{2}{\delta}}}}
{2\widetilde \Delta_{i^*,j} + L_{i^*,j}}}^\frac{2}{3} $ times.
\end{lemma}

Given our bound on the sample size of any sub-optimal arm, we can now bound the overall instance dependent regret. (Recall that the expected rewards are increasing, so we can use $T$ to bound the rewards.)

\begin{theorem}
Algorithm R-ed-AE with $\delta = \frac{1}{2\Phi K T^2}$ guarantees regret 
\eq{
\Re  \leq \sum_{j\in \K\setminus \{i^*\}} \sq*{\frac{  16 \sqrt{\ln\b{4\Phi K T^2 }}}{2\widetilde \Delta_{i^*
,j} + L_{i^*,j}}}^\frac{2}{3} \Phi + 1
.}
\end{theorem}
\begin{proof}
    Under the good event $G$, for every time we sample a sub-optimal arm $i\in \K$,  we increase the regret by at most $\Phi$. The complement event $\bar{G}$ occurs with probability of at most $\frac{1}{T\Phi}$ and the regret in such case is bounded with $T\Phi$. We combine the two and get
    \eq{
\Re \leq \sum_{j\in \K\setminus \{i^*\}} \sq*{\frac{  16 \sqrt{\ln\b{4\Phi K T^2}}}{2\widetilde \Delta_{i^*
,j} + L_{i^*,j}}}^\frac{2}{3} \Phi +2\delta  \Phi K T^2  \leq \sum_{j\in \K\setminus \{i^*\}} \sq*{\frac{  16 \sqrt{\ln\b{4\Phi K T^2}}}{2\widetilde \Delta_{i^*
,j} + L_{i^*,j}}}^\frac{2}{3} \Phi + 1
.}
\end{proof}

\subsection{Arm Elimination with early stopping}

In the case of Rising Rested MAB, there is an issue of playing a sub-optimal arm, even if it is very near to the optimal arm.
For example, if we have two optimal arms, it is not the case that we can mix them, and get an optimal reward. This is due to the fact that we have increasing rewards, and in order to achieve the increase we need to concentrate on playing only one of them. So, unlike the case of regular MAB, where playing any mixture of the two optimal arms is optimal, for Rising Rested MAB we have only two optimal sequences, each consisting of playing the same optimal arm always.

When we consider our instance dependent algorithm, R-ed-AE, we will have an issue if we are left at the end with two near optimal arms. More precisely, we would like to terminate the arm elimination part of the algorithm early, and to force selecting one of the near optimal arms that remain in the set $\K'$.

In addressing such cases, we devised the algorithm HR-ed-AE. The main idea behind the algorithm is using algorithm R-ed-AE as a black box with time horizon of $KM$.
After algorithm R-ed-AE terminates we receive $\K'$, a set of arms which are potentially near-optimal arms. From the set $\K'$ algorithm HR-ed-AE picks an arbitrary arm and plays in the remaining time steps.

\begin{algorithm}
\caption{HR-ed-AE- Halted Rested Arm Elimination}
\begin{algorithmic}[1]
    \State Input: $K$, $T$, $M$,$\delta$
    \State Set $\K' \gets$ R-ed-AE$(K,KM,\delta)$
    \State Select arbitrary $a \in \K'$
    \For{$t \in [KM + 1,T]$}
        \State play $a$
    \EndFor
\end{algorithmic}
\end{algorithm}

The next theorem shows the improvement of HR-ed-AE algorithm. Namely, it has an additional upper bound on the regret, 

\begin{theorem}
Algorithm HR-ed-AE with $\delta = \frac{1}{2\Phi K T^2}$ and $M=T^{4/5}\ln\b{\Phi K T^2}^{1/5}/(\Phi K)^{2/5}$ guarantees regret 
\[\Re \leq O\b{\min\left\{\sum_{j\in \K\setminus \{i^*\}} \sq*{\frac{  16 \sqrt{\ln\b{\Phi K T^2}}}{2\widetilde \Delta_{i^*,j} + L_{i^*,j}}}^\frac{2}{3} \Phi +1\;\; ,\;\; T^\frac{4}{5}(\Phi K)^\frac{3}{5}  \ln\b{\Phi K T^2}^{\frac{1}{5}}\right\}}.\]
\end{theorem}

\begin{proof}
    Recall the definition of the good event $G$ as in Definition 
    \ref{good event G instance dependent AE}. 
    By \Cref{lemma: event G for AE}, event $G$ holds with probability of at least 
    $1-2TK\delta$.
From now on we will assume $G$ holds.

We split the analysis to two cases, according to the arms eliminated during R-ed-AE. If at the end of R-ed-AE there is only a single arm remaining in $\K'$, then, under the event $G$, it is the optimal arm. In this case we will not have any additional regret after R-ed-AE. In this case the regret in the first part of the minimum expression.

From now on assume that two or more arms are in $\K'$, at the end of R-ed-AE. In this case we can bound the regret during R-ed-AE by $KM\Phi$. The main issue is to bound the regret of the selected arm $j\in \K'$ at the end of R-ed-AE.

Since arm $j$ has not been eliminated ($j\in \K'$) in the R-ed-AE, and  $j$ have been sampled at least $M$ times, then,
\eq{
    \sum_{t=1}^{T }\hat{\mu}_{i^*}^M(t)-\hat{\mu}_j^M(t) 
    & < 2\Gamma^{M}(1,T) 
.}

We derive the following lemma to bounds the difference between two values using their estimations.
\begin{lemma}\label{lemma: bound distance between two arms}
    Fix  $x_i,x_j,\hat{x}_i,\hat{x}_j \geq 0$,
    if $|\hat{x}_i - x_i | \leq \gamma$, $|\hat{x}_j - x_j | \leq \gamma$ and $\hat{x}_i-\hat{x}_j \leq 2\gamma$,
    then, $x_i - x_j  \leq 4\gamma$.
\end{lemma}

Given arms $i^*,j\in \K'$, we can utilize \Cref{lemma: bound distance between two arms} and \Cref{cumulative error bound} to derive,
\eq{
    \sum_{t=1}^{T}\mu_{i^*}(t)-\mu_j(t) \leq  4 \Gamma^{M}(1,T)  \leq 2T^2\frac{\sqrt{\ln\b{\frac{2}{\delta}}}}{M^{1.5}}
.}
Therefore, assuming event $G$ holds the regret in the second part is at most 
$O(T^\frac{4}{5}K^\frac{3}{5}  \sqrt{\ln\b{\Phi K T^2}})$.


For the total regret, setting $\delta=1/(2\Phi K T^2)$ and $M=T^{4/5}\ln\b{2/\delta}^{1/5}/(\Phi K)^{2/5}$, we have  that,  
\[\Re \leq MK\Phi +  \frac{  2T^2\sqrt{\ln\b{2/\delta}}}{M^{1.5}} +2\delta \Phi K T^2\leq 3 T^\frac{4}{5}(\Phi K)^\frac{3}{5}  \ln\b{4\Phi K T^2}^{\frac{1}{5}}+  1.\]
\end{proof}

\section{Lower Bound for Rising Rested MAB with Linear Drift}
\label{lower bound section}
In this section we present a lower bound of $\Omega(\min\{T,K^\frac{3}{5}T^\frac{4}{5}\})$ for Rising Rested MAB with Linear Drift.
We concentrate on the case of $K \leq  T^\frac{1}{3}$, since for larger $K$ the lower bound becomes $\Omega(T)$. Our lower bound holds for $\Phi=1$.


The lower bound builds on the fact that we are in a rested MAB model. 
Note that unlike the standard MAB, we can get a regret even if we have two identical arms. If we initially test them for $\tau$ times steps each, then, our regret would be the difference between the cumulative rewards in first $\tau$ time steps compared to the last $\tau$ steps.



Our lower bound construction would set the parameters such that initially we are unable to decide which arm is indeed better, say for the initial $\tau$ steps. Then the regret would be the difference between the  cumulative rewards in first $\tau$ time steps compared to the last $\tau$ steps. More specifically, if one arm has mean rewards of $t/T$, after $t$ times it is used, and the other has $t/T+t/T^{6/5}$, then the KL divergence over the $\tau$ first samples is $\tau^3/T^{12/5}$, which implies that in the first $T^{4/5}$ we will not be able to distinguish between the arms. The regret would be $O(\tau)$, since in the last steps the rewards are almost $1$ and initially the rewards are near zero.

We will now give the details of the lower bound construction.
The core concept of the lower bound is bounding the distance between distributions for specific problem instance. 
We limit our possible profiles to be one of $K+1$ profiles, where each profile gives a complete characterization of the rewards of each arm.
Specifically, we
define the set of profiles as $\I=\{\I_i\}_{i=0}^K$, comprising a set of $K+1$ profiles.
In profile $\I_j$, $j\in \K$, for a given number of samples $t\in [T]$ and arms $i\neq j$, the expected reward of arm $i$ is defined as $\mu_i(t)=\mathcal{N}({t}/{T}-{t K^\frac{3}{5}}/{T^\frac{6}{5}},1)$, where $\mathcal{N}(\mu,\sigma^2)$ is a normal distribution with mean $\mu$ and variance $\sigma^2$. In profile $\I_j$, arm $j$ has expected reward $\mu_j(t)=\mathcal{N}({t}/{T},1)$. So, the optimal arm for profile $\I_j$ is arm $j$.
Additionally, there exists a profile $\I_0$ where for every arm $i \in \K$ and $t\in [T]$ the expected reward is $\mu_i(t)=\mathcal{N}({t}/{T}-{t K^\frac{3}{5}}/{T^\frac{6}{5}},1)$. We will select one of the profiles $\I_j$, $j\in \K$, uniformly at random. 

The following theorem demonstrates that the Rising Rested MAB with Linear Drift setting has a lower bound of $\Omega(T^{\frac{4}{5}}K^\frac{3}{5})$.

\begin{theorem}\label{lower bound Rising Rested MAB with Linear Drift }
    Rising Rested MAB with Linear Drift problem is lower bounded with $\Omega(T^{\frac{4}{5}}K^\frac{3}{5})$, for $K\geq 36$, and $\Omega(T^{\frac{4}{5}})$ otherwise.
\end{theorem}

\noindent{\textit{Proof Sketch:}}
The proof will follows the following general structure. (Detailed proof in Appendix \ref{Apendix: lower bound section}.)

We will consider the first $K\tau/4\geq 9\tau$ steps, where $\tau$ is a parameter we fix later.
When considering profile $\I_0$, at least $3/4$ of the arms are sampled, in expectation, at most $\tau$ times.
We then consider an arm $j$ that has this property.
By the Markov inequality, such an arm $j$ is sampled at most $8\tau$ times with probability at least $7/8$.

Next we bound the $KL$-divergence between $I_j$ and $\I_0$, assuming we sample $j$ at most $\tau$ times. This $KL$- divergence is bounded by $\Theta(\tau^3 K^{6/5}/T^{12/5})$. By Pinsker inequality, this implies that for any event $A$, its probability can differ between $\I_0$ and $\I_j$ by at most $\Theta(\tau^{3/2}K^{3/5}/T^{6/5})$.
For $\tau={T^\frac{4}{5}}/({12 K^\frac{2}{5}})$ the difference in probabilities is at most $1/4$.  

We consider the event that arm $j$ is sampled less than $8\tau$ times. The probability of this event under $\I_0$ is at least $7/8$ so under $\I_j$ the probability is at least $5/8$.

The regret is due to the fact that in $\I_j$, with a constant probability, during the first $K\tau/4$ steps, we sample arm $j$ at most $8\tau$ times. 
Therefore, in at least $K\tau/4 - 8\tau$ steps we use an arm different than $j$. This implies a regret of $\Theta(K\tau)$ when we compare the potential rewards due to arm $j$ with an alternative sub-optimal arm. 

We complete the proof by considering a uniform distribution over the arms, and with probability at least $3/4$ we have a regret of $\Theta(K\tau)$.
$\Box$

\paragraph{Impossibility result when the horizon $T$ is unknown:}
\begin{theorem}
If the horizon $T$ is unknown to the learner, there exists a problem instance where the regret is $\Omega(T)$.
\end{theorem}
\begin{proof}
Consider the follow the following example. A rising rested MAB with linear drift with 2 arms, $a_1$ and $a_2$.
Where for any time step $t \in [T]$ the true values of the arms are, $\mu_{a_1}(t) = 1$ and  $\mu_{a_2}(t) = Lt$, where $L=O(1/T)$. 

For $L<2/T$ the optimal policy is to always play arm $1$, and if $L>2/T$ the optimal policy is to always play arm $2$.

given any online algorithm at time $t=1/L$ we have two events.
\begin{enumerate}
    \item If the algorithm plays arm $1$ less than $1/2L$ times, then we set $T=1/L$. The the algorithm has at most reward of $0.625/L + 1/4$ while the optimal policy (playing always arm $1$) has reward of $1/L$. Hence having a linear regret in $T$.

    \item If the algorithm plays arm $1$ more than $1/2L$ times, then we set $T=3/L$. The algorithm has at most reward of $3.625/L$ while the optimal policy (playing always arm $2$) has reward of $4.5/L$, also having  a linear regret in $T$.
\end{enumerate}
Therefore, for any online algorithm if the horizon is unknown regret is $\Omega(T)$.
\end{proof}

\section{Discussion}
In this work we addressed the rising rested MAB with linear drift problem, i.e., the expectation of the reward function of each arm is linear non-decreasing with respect to the number of times the arm was played.
We bounded the dynamic regret of the problem with both upper and lower bounds,  deriving a tight dynamic regret of $\tilde{\Theta}(T^{4/5}K^{3/5})$. 
In addition we showed an instance dependent regret bound algorithm, which can achieve better regret in some parameters, and in the worst case its regret is still near optimal.

Regarding the rotting rested MAB with linear drift.
The work of \citet{seznec2019rotting} considered the rotting MAB problem, and showed dynamic regret upper bound of $\tilde{O}(\sqrt{KT})$. which solves the rotting rested MAB with linear drift problem with a tight regret bound as well.

It is evident from our results, that in terms of regret, the rising rested MAB with linear drift problem has a higher regret than the rotting rested MAB with linear drift problem or even the stationary stochastic MAB.
Here is an informal intuition why this happens.
In the rising rested MAB with linear drift problem, when we fix an optimal arm, the higher reward values of the optimal arm are obtained when the number of samples are close to $T$.
This implies that each time the learner does not play the optimal arm it might incur a significant regret. For this reason we bound the regret by the number of times we pull the non-optimal arms, and cannot multiply it by a sub-optimality gap, as done in stationary stochastic MAB. This is the major source to the significantly higher regret in our setting, which is inherent, as our lower bound shows.

It is natural to consider a model with both rising and rotting rested MAB with linear drift problem, i.e., the linear slopes can be both positive and negative.
The extension to this case is a fairly straightforward extension of our algorithm. Since the regret would be dominated by the rising arms, the regret bound and analysis would stay almost the same.

\newpage
\bibliography{refs}

\newpage
\appendix
\section{Concentration Lemmas}
\begin{lemma}\label{Hoeffding}Hoeffding's inequality:
Let $X_1,X_2,\ldots,X_m$ independent random variables such for every $i\in[m]$ $X_i\in [0,1]$ and $\mu_i= \E[X_i]$, for $\delta \geq 0$
\[Pr\sq*{\left |\frac{1}{m} \sum_{i\in[m]} \b{X_i-\mu_i}\right| \geq \delta }\leq 2\exp\b{- 2m\delta^2}.\]
\end{lemma}

\begin{lemma}\label{lemma: point estimation  error}
    In Rising Rested MAB with Linear Drift for any arm $i \in \K$ , $m\in \N$ even number, $n'\in \N$ and set of samples $r_i=\{r_i(n'),\ldots,r_i(n'+m)\}$ of arm $i$, with probability of at least $1-\delta$,
    \[ 
    \av{\hat\mu_i^m(n' + \frac{m}{2}) - \mu_i(n' + \frac{m}{2})} \leq 
    \sqrt{ \frac{\ln\b{ \frac{2}{\delta}}}{2m}}
    .\]
\end{lemma}
\begin{proof}
     From the definition we have that $\mu_i(n'+\frac{m}{2}) = \E\sq*{\hat\mu_i^m(n' + \frac{m}{2})} = \E\sq*{\sum_{t=n'}^{n'+m}r_i(n)}$, applying \Cref{Hoeffding} give as,    \[ 
    \av{\hat\mu_i^m(n' + \frac{m}{2}) - \mu_i(n' + \frac{m}{2})} \leq 
    \sqrt{ \frac{\ln\b{ \frac{2}{\delta}}}{2m}}
    .\] 
\end{proof}

\begin{lemma}\label{lemma: bounding estimations }
    In Rising Rested MAB with Linear Drift for any arm $i \in \K$ , $M\in \N$ even number and set of sample 
    $r_i=\{r_i(1),\ldots,r_i(2M)\}$ of arm $i$,
    with probability of at least $1-2\delta$ we have,
    \[\av{\hat\mu_i^{M}\b{\frac{3M}{2}}- \mu_i\b{\frac{3M}{2}}} \leq \sqrt{ \frac{\ln\b{ \frac{2}{\delta}}}{2M}},\] 
    \[\av{\hat\mu_i^{M}\b{\frac{M}{2}}- \mu_i\b{\frac{M}{2}}} \leq 
    \sqrt{ \frac{\ln\b{ \frac{2}{\delta}}}{2M}},\]
    \[\av{ \widehat{L}_i^{2M} - L_i} \leq \frac{ \sqrt{2\ln\b{\frac{2}{\delta}}} } {M^{1.5}}.\]
    
\end{lemma}
\begin{proof}
    Using \Cref{lemma: point estimation  error} we have that with probability of $1-\delta$ each $\av{\hat\mu_i^{M}\b{\frac{3M}{2}}- \mu_i\b{\frac{3M}{2}}} \leq \sqrt{ \frac{\ln\b{ \frac{2}{\delta}}}{2M}}$ and
    $|\hat\mu_i^{M}\b{\frac{M}{2}}- \mu_i\b{\frac{M}{2}}| \leq 
    \sqrt{ \frac{\ln\b{ \frac{2}{\delta}}}{2M}}$,
    using union bound we get that with probability of at least 
    $1-2\delta$ both inequality are holds, and,  
    \[ | \widehat{L}_i^{2M} - L_i| = \frac{|[\hat\mu_i^{M}\b{\frac{3M}{2}} - \hat\mu_i^{M}\b{\frac{M}{2}}] 
            - [ \mu_i\b{\frac{3M}{2}} -\mu_i\b{\frac{M}{2}}]|}{M} ]
     \leq \sqrt{ \frac{2\ln\b{ \frac{2}{\delta}}}{M}}/M.\]

\end{proof}

\begin{lemma}\label{lemma: estimation error of future point}
    In Rising Rested MAB with Linear Drift,
    if $\av{\hat\mu_i^{M}\b{\frac{3M}{2}}- \mu_i\b{\frac{3M}{2}}} \leq \sqrt{ \frac{\ln\b{ \frac{2}{\delta}}}{2M}}$,
    $|\hat\mu_i^{M}\b{\frac{M}{2}}- \mu_i\b{\frac{M}{2}}| \leq 
    \sqrt{ \frac{\ln\b{ \frac{2}{\delta}}}{2M}}$ and $\av{ \widehat{L}_i^{2M} - L_i} \leq \frac{ \sqrt{2\ln\b{\frac{2}{\delta}}} } {M^{1.5}}$
    holds for some arm $i \in \K$, then for any $n\in [T]$,
    \[\av{\psi_i^{2M}(n) - \mu_i(n) }\leq \gamma_n^{2M} .\]
\end{lemma}
\begin{proof}
    We have that:
    \[\mu_i(n) = \frac{\mu_i\b{\frac{3M}{2}} + \mu_i\b{\frac{M}{2}}}{2} + \av{n-M}L_i .\]
    Together with the definition of $\psi_i^{2M}(n)$ we get that, 
    \eq{
    \av{\psi_i^{2M}(n) - \mu_i(n) }
    & = \av{\frac{\hat\mu_i^M(\frac{3M}{2}) + \hat\mu_i^M(\frac{M}{2})}{2} + \av{n-M}  \widehat{L}_i^{2M} - \frac{\mu_i(\frac{3M}{2}) + \mu_i(\frac{M}{2})}{2} -  \av{n-M}L_i  }
    \\ & \leq 
     \sqrt{\frac{\ln\b{\frac{2}{\delta}}}{2M}} 
    + \av{n - M} 
    \frac{\sqrt{2\ln\b{\frac{2}{\delta}}}}
    {M^{1.5}}
    = \gamma_n^{2M}
    .}
\end{proof}

\begin{lemma}\label{lemma: cumulative error estimation }
    In Rising Rested MAB with Linear Drift,
    if $\av{\hat\mu_i^{M}\b{\frac{3M}{2}}- \mu_i\b{\frac{3M}{2}}} \leq \sqrt{ \frac{\ln\b{ \frac{2}{\delta}}}{2M}}$,
    $|\hat\mu_i^{M}\b{\frac{M}{2}}- \mu_i\b{\frac{M}{2}}| \leq 
    \sqrt{ \frac{\ln\b{ \frac{2}{\delta}}}{2M}}$ and $\av{ \widehat{L}_i^{2M} - L_i} \leq \frac{ \sqrt{2\ln\b{\frac{2}{\delta}}} } {M^{1.5}}$
    holds for some arm $i \in \K$, then 
    \[\av{\hat{s}_i^{2M}(n_1,n_2) - s_i(n_1,n_2)}
    \leq \Gamma^{2M}(n_1,n_2) .\]
\end{lemma}
\begin{proof}
    Using Lemma  \ref{lemma: estimation error of future point} we have that $\av{\psi_i^{2M}(n) - \mu_i(n) }\leq \gamma_n^{2M}$ and we get that, 
    \[\av{\hat{s}_i^{2M}(n_1,n_2) - s_i(n_1,n_2)} 
    \leq \sum_{n=n_1}^{n_2} \av{\psi_i^{2M}(n) - \mu_i(n)}
    \leq \sum_{n=n_1}^{n_2} \gamma_n^{2M} = \Gamma^{2M}(n_1,n_2)
    .\]
\end{proof}

\begin{lemma}\label{cumulative error bound}
    For any $n_1,n_2\in [T]$,
    \[\Gamma^{2M}(n_1,n_2) \leq \Gamma^{2M}(1,T)\leq T^2\frac{\sqrt{\ln\b{\frac{2}{\delta}}}}{\sqrt{2}M^{1.5}}.\]
\end{lemma}
\begin{proof}
   \eq{
     \Gamma^{2M}(n_1,n_2) &\leq \Gamma^{2M}(1,T) = 
     \sum_{n=1}^{T} \sqrt{\frac{\ln\b{\frac{2}{\delta}}}{2M}} 
    + |n - M| \frac{\sqrt{2\ln\b{\frac{2}{\delta}}}}{M^{1.5}}
    \\ & = 
    2M\sq*{\sqrt{\frac{\ln(\frac{2}{\delta})}{2M}} 
    + \frac{M+1}{2}  \frac{\sqrt{2\ln(\frac{2}{\delta})}}{M^{1.5}}}
    +(T-2M)\sq*{\sqrt{\frac{\ln(\frac{2}{\delta})}{2M}} 
    + \b{\frac{T+1}{2}} \frac{\sqrt{2\ln(\frac{2}{\delta})}}{M^{1.5}}}
        \\ & = 
    (2M^2+M) \frac{\sqrt{2\ln\b{\frac{2}{\delta}}}}{M^{1.5}}
    +(T-2M)\b{\frac{T+M+1}{2}} \frac{\sqrt{2\ln\b{\frac{2}{\delta}}}}{M^{1.5}}
    \\ & \leq T^2\frac{\sqrt{\ln\b{\frac{2}{\delta}}}}{\sqrt{2}M^{1.5}}
    .}
\end{proof}

\section{Proof from \autoref{section: first alg}}
\begin{manuallemma}{\ref{lemma:G}}
    The probability of the good even $G$ is at lest $1-2\delta K$. 
\end{manuallemma}
\begin{proof}
From \Cref{lemma: bounding estimations }, for each arm $i \in \K$, using the $2M$ samples of arm $i$, we have that with probability of at least $1-2\delta$ :
    \[\av{\hat\mu_i^{M}\b{\frac{M}{2}} - \mu_i\b{\frac{M}{2}}}\leq \sqrt{ \frac{\ln\b{ \frac{2}{\delta}}}{2M}},\] 
    \[ \av{\hat\mu_i^{M}\b{\frac{3M}{2}} - \mu_i\b{\frac{3M}{2}}|}\leq \sqrt{ \frac{\ln\b{ \frac{2}{\delta}}}{2M}},\]
    \[\av{ \widehat{L}_i^{2M} - L_i|}\leq \frac{ \sqrt{2\ln\b{\frac{2}{\delta}}} } {m^{1.5}}.\]
    Using union bound over all $K$ arms we get that $G$ holds with probability of at least $1-2\delta K$.
\end{proof}

\begin{claim}
    \[\hat{s}_i^{2M}(n_1,n_2) = (n_2-n_1+1)\sq*{\b{\hat\mu_i^M\b{\frac{3M}{2}} + \hat\mu_i^M\b{\frac{M}{2}}}/2 
        + \b{\frac{n_2+n_1}{2} - M} \widehat{L}_i^{2M}}.\]
\end{claim}
\begin{proof}
\eq{
\hat{s}_i^{2M}(n_1,n_2) & =\sum_{t=n_1}^{n_2}\psi^{2M}_i(n) 
\\ & = \sum_{t=n_1}^{n_2} \b{\hat\mu_i^M\b{\frac{3M}{2}} + \hat\mu_i^M\b{\frac{M}{2}}}/2 
        + (n - M) \widehat{L}_i^{2M}
\\ & = (n_2-n_1+1)\sq*{\b{\hat\mu_i^M\b{\frac{3M}{2}} + \hat\mu_i^M\b{\frac{M}{2}}}/2 
        + \b{\frac{n_2+n_1}{2} - M} \widehat{L}_i^{2M}}
.}    
\end{proof}

\section{Proof from \autoref{instance dependent section}}

\begin{manuallemma}{\ref{lemma: event G for AE}}
    The probability of the good even $G$ is at lest $1-2TK\delta$.
\end{manuallemma}
\begin{proof}\label{proof for Lemma event G for AE}
From \Cref{lemma: bounding estimations } we get that for any arm $i \in \K$ and $m\in [T]$ with probability of at least $1-2\delta$: $\av{\hat\mu_i^{\frac{m}{2}}(\frac{m}{4}) - \mu_i(\frac{m}{4})} \leq \sqrt{\frac{\ln\b{\frac{2}{\delta}}}{m}}$, 
 $\av{\hat\mu_i^{\frac{m}{2}}(\frac{3m}{4})-\mu_i(\frac{3m}{4})}\leq\sqrt{\frac{\ln\b{\frac{2}{\delta}}}{m}}$ 
 and $\av{ \widehat{L}_i^{m}-L_i}\leq \frac{ 2\sqrt{\ln\b{\frac{2}{\delta}}} } {m^{1.5}}$.
using union bound over $T$ and $K$ we have that event $G$ is occur with probability of at least $1-2TK\delta$.
\end{proof}

\begin{manuallemma}{\ref{Lemma: best ar never eliminated}}
    Under $G$ we never eliminate the optimal arm $i^*$.
\end{manuallemma}
\begin{proof}
    Assuming event $G$ hold and $i^*\notin \K'$.
    We conclude that after $N_{i^*}$ samples of arm $i^*$ the algorithm R-ed-AE eliminated  arm $i^*$ using some arm $j\in \K'$, where $N_{i^*}=N_j$.
    Thus the elimination term was satisfy, namely,
    \[ \hat{s}_j^{N_{i^*}}(1,T) - \hat{s}_{i^*}^{N_{i^*}}(1,T)  > 2 \Gamma^{N_{i^*}}(1,T)
    .\]
    In addition, from event $G$ we have that for any arm $i\in \K'$ and $t \in [T]$, 
     \eq{
     &|\mu_{i}(t) - \psi^{N_{i^*}}_i(t)| \leq \sqrt{\frac{\ln\b{\frac{2}{\delta}}}{N_{i^*}}} + \left | n-\frac{M}{2} \right |\frac{2\sqrt{\ln\b{\frac{2}{\delta}}}}{N_{i^*}}  = \gamma^{N_{i^*}}_t}
     Therefore, 
     \eq{
     |s_i(1,T) - \hat{s}_i^{N_{i^*}}(1,T)|\leq \Gamma^{N_{i^*}}(1,T)
    .}
    From both inequalities we have,
        \eq{
        \hat{s}_j^{N_{i^*}}(1,T) - \hat{s}_{i^*}^{N_{i^*}}(1,T)  &>  s_{i^*}(1,T) - \hat{s}_{i^*}^{N_{i^*}}(1,T) + \hat{s}_j^{N_{i^*}}(1,T) - s_j(1,T)}
        Hence,
        \eq{
            s_j(1,T) &>  s_{i^*}(1,T) 
    .}
    In contradiction to the optimality of arm $i^*$.
\end{proof}

\begin{manuallemma}{\ref{elimination Lemma R-ed-AE}}
Algorithm R-ed-AE samples a sub-optimal arm $j$ at most
$\sq*{\frac{  16 \sqrt{\ln\b{\frac{2}{\delta}}}}
{2\widetilde \Delta_{i^*,j} + L_{i^*,j}}}^\frac{2}{3} $ times.
\end{manuallemma}
\begin{proof}
Consider the elimination of arm $j$ due to arm $i$. This implies that so far we have played each of $i$ and $j$ for $N_j$ times each.
Then, using that event $G$ holds and using \Cref{lemma: cumulative error estimation } for $N_j$, we have the following for the rewards of each of $i$ and $j$.
When we eliminate arm $j$ (by $i$) we have
\begin{equation} \label{N_j inequality1}
\hat{s}_i^{N_j}(1,T)-\hat{s}_j^{N_j}(1,T)   =
    \sum_{n=1}^{T}\hat{\mu}_i(t,N_j)-\hat{\mu}_j(n,N_j) \geq  \sum_{n=1}^{T} 2 \gamma_n^{N_j} = 2 \Gamma^{N_j}.
\end{equation}

Due to the definition of $\gamma_n^{N_j}$, we have from \Cref{lemma: bound distance between two arms} regrading the true rewards,
\begin{equation}\label{N_j inequality2}
\sum_{n=1}^{T}\mu_{i}(n)-\mu_j(n) \geq 
\sum_{n=1}^{T}\hat{\mu}_i(n,N_j)-\hat{\mu}_j(n,N_j) +  \sum_{n=1}^{T} 2 \gamma_n^{N_j}  \geq  \sum_{n=1}^{T} 4 \gamma_n^{N_j} = 4 \Gamma^{N_j}(1,T).
\end{equation}
   
Since, by definition of $i^*$, we have that 
$\sum_{t=1}^{T}\mu_{i^*}(t)\geq \sum_{t=1}^{T}\mu_{i}(t)$, and $N_j=M/2$, from \Cref{cumulative error bound}
\[
\Gamma^{N_j}(1,T) \leq T^2\frac{\sqrt{\ln\b{\frac{2}{\delta}}}}{\sqrt{2}M^{1.5}}\leq 2T(T+1)\frac{\sqrt{\ln\b{\frac{2}{\delta}}}}{N_j^{1.5}}.
\]
Consider,
\[
    N_j
    \geq 
        \sq*{\frac{16 (T+1)  \sqrt{\ln\b{\frac{2}{\delta}}}}{2\Delta_{i^*,j} + (T+1)L_{i^*,j}}}^\frac{2}{3}
.\]
Then, 
\eq{
 \Rightarrow & 
    N_j^{1.5}
    \geq 
        \frac{8 (T+1)  \sqrt{\ln\b{\frac{2}{\delta}}}}{\Delta_{i^*,j} + \frac{T+1}{2}L_{i^*,j}}
\\ \Rightarrow & 
   \Delta_{i^*,j} + \frac{T+1}{2}L_{i^*,j} 
    \geq 
     8 (T+1)   \frac{  \sqrt{\ln\b{\frac{2}{\delta}}}}{N_j^{1.5}}
\\ \Rightarrow &
    T(\Delta_{i^*,j} + \frac{T+1}{2}L_{i^*,j}) 
    \geq 
    8 T(T+1) \frac{  \sqrt{\ln\b{\frac{2}{\delta}}}}{N_j^{1.5}}\\
&
    \sum_{t=1}^{T}\mu_{i^*}(t)-\mu_j(t) \geq  
    8T(T+1)\frac{\sqrt{\ln\b{\frac{2}{\delta}}}}{N_j^{1.5}}
    .}
This implies that after such $N_j$ arm $i^*$ would have eliminated arm $j$. If arm $i^*$ eliminated arm $j$, this can only happen earlier.

Rewriting,  the expression for $N_j$,
\eq {
   \sq*{\frac{  16(T+1)\sqrt{\ln\b{\frac{2}{\delta}}}}{2\Delta_{i^*,j} + (T+1)L_{i^*,j}}}^\frac{2}{3}
   \leq 
   \sq*{\frac{  16\sqrt{\ln\b{\frac{2}{\delta}}}}{2\widetilde \Delta_{i^*,j} + L_{i^*,j}}}^\frac{2}{3} 
.}
Then, for values of $N_j\geq \sq*{\frac{  16\sqrt{\ln\b{\frac{2}{\delta}}}}{2\widetilde \Delta_{i^*,j} + L_{i^*,j}}}^\frac{2}{3}$ we are guaranteed that arm $j$ was already eliminated.

Therefore, the condition for elimination is hold for $\sq*{\frac{  16\sqrt{\ln\b{\frac{2}{\delta}}}}{2\widetilde \Delta_{i^*,j} + L_{i^*,j}}}^\frac{2}{3} $.
\end{proof}

\begin{manuallemma}{\ref{lemma: bound distance between two arms}}
       Fix  $x_i,x_j,\hat{x}_i,\hat{x}_j \geq 0$,
    if $|\hat{x}_i - x_i | \leq \gamma$, $|\hat{x}_j - x_j | \leq \gamma$ and $\hat{x}_i-\hat{x}_j \leq 2\gamma$,
    then, $x_i - x_j  \leq 4\gamma$.
\end{manuallemma}
\begin{proof}
$x_i - x_j  \leq \hat{x}_i - \hat{x}_j + 2 \gamma \leq 4\gamma$.
\end{proof}

\section{Dynamic vs static regret}

\begin{definition} 
\begin{enumerate}$\:$
    \item Let $V$ be a set of vectors, where $V=\cb{v\in \{\N\cup \{0\} \}^K |\; \norm{v}_1=T}$.

    \item For $i\in \K$, let $e_i\in \R^K$ be the unit vector with $1$ at the i'th entry and 0 at the rest.
    


        
\end{enumerate}
\end{definition}

    \begin{lemma}\label{moving wight - lemma}
        For any two linear monotonic increasing functions $f(x)=ax+b$ and $g(x)+cx+d$
        and any $\alpha,\beta\in \N\cup\{0\}$,
       assuming w.l.o.g. 
       $\sum_{x=1}^{\alpha+\beta}f(x) \geq \sum_{x=1}^{\alpha+\beta}g(x)$,
       then the follow holds, 
       \eq{
        \sum_{x=1}^{\alpha+\beta}f(x)
       & \geq \sum_{x=1}^{\alpha}f(x) +\sum_{x=1}^{\beta}g(x)
       .}
    \end{lemma}
   \begin{proof}
   Note that,
   \begin{align}\label{eq:f+g}
       \sum_{x=1}^{\alpha+\beta}f(x)
       & \geq \sum_{x=1}^{\alpha}f(x) +\sum_{x=1}^{\beta}g(x)
        \qquad \Longleftrightarrow  \qquad
        \sum_{x=\alpha+1}^{\alpha+\beta}f(x)
        \geq \sum_{x=1}^{\beta}g(x)
       .
\end{align}
   \begin{enumerate}[label=\arabic*., ref=\arabic*]
       \item If for any $x\in[\alpha+\beta]$ we have $f(x)\geq g(x)$, we get that,
       \[\sum_{x=\alpha+1}^{\alpha+\beta}f(x)\geq \sum_{x=1}^{\beta}f(x)
       \geq \sum_{x=1}^{\beta}g(x),\]
       and we are done.
       \item Else, $f$ and $g$ have intersection point $z=\frac{d-b}{a-c}$.
       %
           In the case that for any $x\in[\left \lfloor z \right \rfloor]$ we have $f(x)\geq g(x)$ and for any $x\in[\left \lceil z\right \rceil, \alpha+\beta]$ we have $f(x)\leq g(x)$.
           
           \begin{enumerate}[label*=\arabic*., ref=\theenumi.\arabic*]
                \item If $\beta \leq z$, 
                    \[\sum_{x=\alpha+1}^{\alpha+\beta}f(x)\geq \sum_{x=1}^{\beta}f(x)\geq \sum_{x=1}^{\beta}g(x),\]
            and we are done.

                \item If $\beta \geq z$,  
                    \eq{
                    & \sum_{x=\beta+1}^{\alpha+\beta}g(x)
                    \geq \sum_{x=\beta+1}^{\alpha+\beta}f(x)
                    \geq \sum_{x=1}^{\alpha}f(x)}
            since $f(x)$ is non-decreasing. Hence,
                    \eq{ 
                    \sum_{x=1}^{\alpha+\beta}f(x)
                        \geq  \sum_{x=1}^{\alpha+\beta}g(x)
                        = \sum_{x=\beta+1}^{\alpha+\beta}g(x) +\sum_{x=1}^{\beta}g(x)
                         \geq \sum_{x=1}^{\alpha}f(x) +\sum_{x=1}^{\beta}g(x)
                   ,}
            and we are done.

              \item Otherwise for $x\in[\left \lfloor z \right \rfloor]$ we have $f(x)\leq g(x)$ and for $x\in[\left \lceil z\right \rceil, \alpha+\beta]$ we have $f(x)\geq g(x)$.
               \begin{enumerate}[label*=\arabic*., ref=\theenumi.\arabic*]
                    \item If $\alpha \leq z$,
                        \eq{
                        & \sum_{x=\beta+1}^{\alpha+\beta}g(x)
                        \geq \sum_{x=1}^{\alpha}g(x)
                        \geq \sum_{x=1}^{\alpha}f(x)
                        \\ \Rightarrow & 
                            \sum_{x=1}^{\alpha+\beta}f(x)
                            \geq  \sum_{x=1}^{\alpha+\beta}g(x)
                            = \sum_{x=\beta+1}^{\alpha+\beta}g(x) +\sum_{x=1}^{\beta}g(x)
                             \geq \sum_{x=1}^{\alpha}f(x) +\sum_{x=1}^{\beta}g(x)
                        .}
    
                    \item If $\alpha \geq z$,  
                        \eq{
                        & \sum_{x=\alpha+1}^{\alpha+\beta}f(x)
                        \geq \sum_{x=\alpha+1}^{\alpha+\beta}g(x)
                        \geq \sum_{x=1}^{\beta}g(x)
                        .}
       \end{enumerate}
   \end{enumerate}
\end{enumerate}      
\end{proof}



\begin{lemma} \label{vector representetion}
    In the Rested MAB problem with horizon $T$, a trajectory of a policy can be represented as vector $v\in \{\N\cup \{0\} \}^K$ with $\norm{v}_1=T$.
\end{lemma} 
\begin{proof}
    The return of any trajectory  dependent only on the number of time the policy sampled each arm.
    Fix a trajectory $a=\{a_1,a_2,\ldots,a_T\}$ lets define the vector $v\in \R^K$ as follow: for any $i\in \K$, the i'th entry  $v_i=\sum_{t=1}^{T}
     \mathbbm{1} (a_t=i) = N_i(T)$, 
     and we get that \[\sum_{t=1}^{T}\mu_{t}(N_t(a_t))= \sum_{i\in \K}^{}\sum_{t=1}^{v_i}\mu_{i}(t).\]
\end{proof}

Note: thanks to \Cref{vector representetion} From now on when we talk about a vector in relation to the policy, we will mean the vector representing the policy run and vice versa.  

 \begin{theorem}\label{theorem: Rise-Res-LinD-Opt-Policy}
    The optimal policy for Rising Rested MAB with Linear Drift is playing always arm $i^*$, i.e.,
    for any $v \in V$:
    \[
    \sum_{i\in \K}^{} \sum_{t=1}^{d_i} \mu_i(t)
    \leq
    \underset{i\in \K}{\max}\sum_{t=1}^{T}\mu_i(t)
    .\]
\end{theorem}
\begin{proof}
    define $i^* = \underset{i\in \K}{\argmax}\sum_{t=1}^{T}\mu_i(t)$ as the best arm.
    
    For any vector $v\in V\setminus \{e_i \cdot T\}_{i\in \K}$,
    from \Cref{moving wight - lemma} we know that there exist $j,l\in \K$ when $l \neq j$ and $v_j,v_l\geq 1$ such: 
    \eq{
    & \sum_{n=1}^{v_j+v_l}\mu_l(n)
        \geq \sum_{n=1}^{v_j}\mu_j(n) +\sum_{n=1}^{v_l}\mu_l(n)
    \\ \Rightarrow & \sum_{i\in \K}^{} \sum_{n=1}^{v_i}\mu_i(n)\leq 
        \sum_{i\in \K\setminus \{j\}}^{} \sum_{n=1}^{v_i}\mu_i(n)+\sum_{n=v_l+1}^{v_l+v_j}\mu_l(n)
    .}
    
    Hence, for any vector $v\in V\setminus \{e_i \cdot T\}_{i\in \K}$, if we apply \Cref{moving wight - lemma} at most $K$ times we will get that there exists $j\in \K$ s.t.:
    \[\sum_{i\in \K}^{}\sum_{n=1}^{v_i}\mu_i(n)\leq
    \sum_{n=1}^{T} \mu_{j} (n).\]
    
    Since $i^* = \underset{i\in \K}{\argmax}\sum_{n=1}^{T}\mu_i(n)$ we get that, 
    \[\sum_{n=1}^{T}\mu_{i^*}(n)\geq\underset{v\in V}{\max} \cb{\sum_{i\in \K}^{}\sum_{n=1}^{v_i}\mu_i(n)}.\]
\end{proof}

\section{Proofs for \autoref{lower bound section}}\label{Apendix: lower bound section}

\begin{definition}\label{Definition: lower bound}
    \begin{itemize}$\:$
        \item Distributions:
    \begin{equation*}
    \begin{split}
    &q_t\sim \mathcal{N}(\frac{t}{T}-\frac{t K^\frac{3}{5}}{T^\frac{6}{5}},1)
    \text{,\;\;\;}
    Q_m=q_1\times \ldots\times q_m. 
    \\&
    p_t\sim \mathcal{N}(\frac{t}{T},1) 
    \text{,\;\;\;\;\;\;\;\;\;\;\;\;\;\;\;}
    P_m=p_1\times \ldots\times p_m . 
    \end{split}
    \end{equation*}

    \item 
    We define a set  $\I$ of  $K+1$ profiles of rested bandits problems with linear drift, where for each $j\in \K$ $\forall t \in T $ the profile $\I_j$ is:
    \[\I_j=\begin{cases}
    \mu_i(t)= \E[q_t] & \text{ if } i\neq j \\
    \mu_i(t)= \E[p_t] & \text{ if } i=j
    \end{cases}.\]

    We also define a special profile $\I_0$:
    \[\I_0=\begin{cases}
    \mu_i(t)= \E[q_t] 
    \end{cases}.\]
    
    \end{itemize}
\end{definition}

\begin{manualtheorem}{\ref{lower bound Rising Rested MAB with Linear Drift }} 
    Rising Rested MAB with Linear Drift problem is lower bounded with $\Omega(T^{\frac{4}{5}}K^\frac{3}{5})$, for $K\geq 36$, and $\Omega(T^{\frac{4}{5}})$ otherwise.
\end{manualtheorem}
\begin{proof}
For $K\geq 36$
Fix an arm $j$ satisfying the follow property: 
\begin{equation}\label{sufficient j}
  \E_0[N_j(\frac{1}{4}K\tau)] \leq \tau.  
\end{equation}

Since there are at least $\frac{3}{4}K$ such arms at the $\I_0$ instance, this implies that if we select arm $j$ uniformly at random we have $3/4$ probability that it holds, i.e., $\Pr_{j\in \K}[\E[N_j(\frac{1}{4}K\tau)]\leq \tau|\;\I_0]\geq \frac{3}{4}$.
If $\E_0[N_j(\frac{1}{4}K\tau)] \leq \tau$, then by Markov inequality we have, \[\Pr[N_j(\frac{1}{4}K\tau)] \leq 8\tau|\I_0]\geq 7/8.\]

We will now refine our definition of the sample space. For each arm $a$,
define the $t$-round sample space $\Omega^t_a = [0,1]^t$,
where each outcome corresponds to a particular realization of the tuple
$(r_a(s):s\in [t])$.
Then, the partial first $\frac{1}{4}K\tau$ sample space we considered before can be expressed as
$\Omega = \prod_{a \in \K} \Omega^{\frac{1}{4}K\tau}_a$.

We consider a “reduced” sample space in which arm $j$ played at most $8\tau$ times:
\[\Omega^*_j = \Omega^{8\tau}_j \times \prod_{a\neq j}\Omega^{\frac{1}{4}K\tau}_a.\]
We will be interested in the event $A=\{N_j(\frac{1}{4}K\tau)] \leq 8\tau\}$. Note that the event $A$ is determined by $\Omega^*_j$.

For each profile $\I_j$ 
, we define distribution $Z_j^*$ on $\Omega^*_j$ as follows: 
\[Z^*_j (A) = \Pr[A|\I_j] \text{ for each } A\subset \Omega^*_j.\] 

We want to bound the difference between the distributions $Z^*_0(A)$ and $Z^*_j(A)$.
Using \Cref{Lemma: distribution distance}, we have that,
\[
|Z^*_0(A) - Z^*_j(A)| \leq 10\frac{\tau^{1.5} K^\frac{3}{5}}{ T^\frac{6}{5}}
.\]
%
For $\tau=\frac{T^\frac{4}{5}}{12 K^\frac{2}{5}}$  
we obtain that $|Z^*_0(A) - Z^*_j(A)| \leq \frac{1}{4}$.
Therefore, for arm $j$ for which property (\ref{sufficient j}) holds, we attain
\[Pr[N_j(\frac{1}{4}K\tau) \geq 8\tau|\;\I_j] \leq \frac{1}{4}+ Pr[N_j(\frac{1}{4}K\tau) \geq 8\tau|\;\I_0] \leq 3/8.\]
This implies that $Pr[N_j(\frac{1}{4}K\tau) \leq 8\tau|\;\I_j] \geq 5/8$.

Again, assuming that for arm $j$  property (\ref{sufficient j}) holds over instance $\I_0$, when playing "full" game over instance $\I_j$, with probability at least $5/8$ arm $j$ will be sample at most $T-(\frac{K}{4}-8)\tau$ times, i.e.,
$N_j(T)|\;I_j \leq T-(\frac{K}{4}-8)\tau$.

From \Cref{Lemma: regret for instance lower bound}, since with probability at least $5/8$ we have $N_j(T)\leq T-(\frac{K}{4}-8)\tau$, then we have that, 
$\E _j[\Re] \geq \frac{5}{8}\frac{K^\frac{3}{5}T^\frac{4}{5}}{40}$. Therefore, for a uniform random profile $\I_j$ we have an expected of $\E[\Re] \geq \frac{K^\frac{3}{5}T^\frac{4}{5}}{64}$.

For $K=2$. 
Let the two arms be $a_q$ and $a_p$ when $r_p(t)\sim \mathcal{N}(\frac{t}{T},1)$ and $r_q(t)\sim \mathcal{N}(\frac{t}{T}-\frac{t}{\sqrt{2}T^\frac{6}{5}},1)$.
Applying Pinsker's inequality and KL-divergent we get that,
    \eq{
    & 2\norm{P_\tau - Q_\tau}_1^2 KL(P_\tau || Q_\tau)
    \leq \sum_{t=1}^{\tau} \b{\frac{t}{\sqrt{2}T^\frac{6}{5}}}^2 
    \leq \frac{\tau^3 }{2T^\frac{12}{5}}      
    \\ & 
    \Rightarrow \norm{P_\tau - Q_\tau}_1 \leq \frac{\tau^{1.5}}{2T^\frac{6}{5}}
    .}
Therefore, for $\tau \leq T^\frac{4}{5}$ we have that $ \norm{P_\tau - Q_\tau}_1 \leq \frac{1}{2}$, meaning that the two distributions are indistinguishable if both sampled less than $\tau$ times.
Hence, to distinguish between the arms the player have to sample one arm at least $T^\frac{4}{5}$ times.
The probability of sampling the bad arm $T^\frac{4}{5}$ times is $\frac{1}{2}$. Giving a regret of,
\eq{
    \Re &\leq  0.5 \sum_{t=1}^{T^\frac{4}{5}}
    \frac{T-t+1}{T}
    - \frac{t}{T}+\frac{t}{\sqrt{2}T^\frac{6}{5}}
    \\ & \geq 
    0.5T^\frac{4}{5}
    - \frac{T^\frac{8}{5}+T^\frac{4}{5}}{2}\frac{1}{T}
        \\ & =
    0.5(T^\frac{4}{5}
    - T^\frac{3}{5}+T^{-\frac{1}{5}})
        \\ & \geq  0.25T^\frac{4}{5}
.}

\end{proof}

\begin{lemma}\label{Lemma: distribution distance}
For every instance $j\in \K$,
for any event $A\subset \Omega^*$, we have
$
|Z^*_0(A) - Z^*_j(A)|
\leq 10 \frac{\tau^{1.5} K^\frac{3}{5}}{T^\frac{6}{5}}$, i.e., $\|Z^*_0-Z^*_j\|_1\leq 10\frac{\tau^{1.5} K^\frac{3}{5}}{T^\frac{6}{5}} $.
\end{lemma}
\begin{proof}
Applying Pinsker's inequality and KL-divergent we will have for any event $A\subset \Omega^*$:

\eq{
2(Z^*_0(A) - Z^*_j(A) )^2                                     
& \leq KL(Z^*_0||Z^*_j)                                       
\\&= \sum_{a\in \K} \sum_{t=1}^{N_a(\frac{1}{4} K \tau)}KL(Z^{a,t}_0||Z^{a,t}_j)     
\\& = \sum_{a \neq j} \sum_{t=1}^{N_a(\frac{1}{4} K \tau)}KL(Z^{a,t}_0||Z^{a,t}_j) +                  \sum_{t=1}^{N_j(\frac{1}{4} K \tau)}KL(Z^{j,t}_0||Z^{j,t}_j)                 
\\&\leq \sum_{t=1}^{8\tau} \b{\frac{t K^\frac{3}{5}}{T^\frac{6}{5}}}^2                                                                 
\leq 180\frac{\tau^3 K^\frac{6}{5}}{T^\frac{12}{5}}              
.}
In the inequality we use the assumption $N_j(\frac{1}{4} K \tau) \leq 8\tau$ and that for any $a\neq j$ $Z^{a,t}_0$ and $Z^{a,t}_j$ are the same. Therefore,
\[ |Z^*_0(A) - Z^*_j(A)|                  
\leq 10\frac{\tau^{1.5} K^\frac{3}{5}}{T^\frac{6}{5}}.\]
\end{proof}

\begin{definition}\label{Definitions: for egret for instance lower bound proof}  $\:$
    \begin{itemize}
        \item Let the vector $v=\{ v_i \} _ {i \in \K}$, where $v_i=N_i(T)$ represent the number of times arm $i$ was sampled, note that $\norm{v}_1 = T$.
        
        \item Let the vector $v'$ be the vector $v$ where we zero the entry for arm $j$.
        
        \item $V$ a set of vectors $v$ such that the entries 
        $v_j\leq T-(\frac{K}{4}-8)\tau$. i.e.,
        $V=\{v\in  \N^k : 
        \norm{v}_1 = T, v_j 
        \leq  T-(\frac{K}{4}-8)\tau\}$
    \end{itemize}
\end{definition}

\begin{lemma}\label{Lemma: regret for instance lower bound}
    When using instance $\I_j$ assuming $N_j(T)\leq T-(\frac{K}{4}-8)\tau$ the regret is lower bounded with $\frac{K^\frac{3}{5}T^\frac{4}{5}}{40}$. i.e.,
    $\E _j^*[\Re] \geq \frac{K^\frac{3}{5}T^\frac{4}{5}}{40}$.
\end{lemma}

\begin{proof}
In this proof we will use Definition \ref{Definitions: for egret for instance lower bound proof}.
The regret when using instance $\I_j$ assuming $E[N_j(T)]\leq T-(\frac{K}{4}-8)\tau$.
Formally, we define $\E _j^*[\Re]=\E [\Re|\;\; \I_j, E[N_j(T)]\leq T-(\frac{K}{4}-8)\tau]$.
\eq{
\E_j^*[\Re]   & \geq                                        
\underset{v\in V}{\min}\cb{
\sum_{t=1}^{T} \frac{t}{T}                                        
- \b{ \sum_{i\in \K \setminus  \cb{j}}                           
\sum_{t=1}^{v_i}\sq*{\frac{t}{T}-\frac{tK^\frac{3}{5}}{T^\frac{6}{5}}}   
+ \sum_{t=1}^{v_j} \frac{t}{T}}}                       
\\ & =   \underset{0 \leq \beta \leq T-(\frac{K}{4}-8)\tau}{\min}\cb{     
\sum_{t=1}^{T} \frac{t}{T}                          
- \b{  \sum_{t=1}^{T-\beta }\sq*{\frac{t}{T}-\frac{tK^\frac{3}{5}}{T^\frac{6}{5}}}   
+ \sum_{t=1}^{\beta } \frac{t}{T}}      }                                                
\\ & =  \underset{0 \leq \beta \leq T-(\frac{K}{4}-8)\tau}{\min}\cb{      
\sum_{t=\beta +1}^{T} \frac{t}{T}                            
-  \sum_{t=1}^{T-\beta }\frac{t}{T}                          
+ \sum_{t=1}^{T-\beta }\frac{tK^\frac{3}{5}}{T^\frac{6}{5}}} 
\\ & = \underset{0 \leq \beta \leq T-(\frac{K}{4}-8)\tau}{\min}\cb{        
\beta                                               
-\frac{\beta^2}{T}                                  
+ \frac{K^\frac{3}{5} T^\frac{4}{5}}{2}             
+ \frac{K^\frac{3}{5} \beta^2}{2T^\frac{6}{5}}      
+ \frac{K^\frac{3}{5} \beta}{T^\frac{1}{5}}         
+\frac{K^\frac{3}{5}}{2T^\frac{1}{5}}               
- \frac{K^\frac{3}{5}\beta}{2T^\frac{6}{5}}         
}      
.}
In the first equality we use that 
$\sum_{i \neq j}^{}
\sum_{t=1}^{v_i}\sq*{\frac{t}{T}-\frac{t K^\frac{3}{5}}{T^\frac{6}{5}}} 
\leq 
\sum_{t=1}^{\norm{\acute{v}}_1}\sq*{\frac{t}{T}-\frac{t K^\frac{3}{5}}{T^\frac{6}{5}}}$.

Given $f(\beta) = \beta                            
-\frac{\beta^2}{T}                                  
+ \frac{K^\frac{3}{5} T^\frac{4}{5}}{2}             
+ \frac{K^\frac{3}{5} \beta^2}{2T^\frac{6}{5}}      
+ \frac{K^\frac{3}{5} \beta}{T^\frac{1}{5}}         
+\frac{K^\frac{3}{5}}{2T^\frac{1}{5}}               
- \frac{K^\frac{3}{5}\beta}{2T^\frac{6}{5}}         
$
, lets calculate the first and second moments of $f$:
\[
f'(\beta) = 
1                                              
-\frac{2\beta}{T}                                  
+ \frac{K^\frac{3}{5} \beta}{T^\frac{6}{5}}      
+ \frac{K^\frac{3}{5} }{T^\frac{1}{5}}          
- \frac{K^\frac{3}{5}}{2T^\frac{6}{5}}, 
\]

\[
f''(\beta) =     
-\frac{2}{T}                               
+ \frac{K^\frac{3}{5} }{T^\frac{6}{5}} 
\leq -\frac{2}{T} + \frac{T^{\frac{1}{5}} }{T^\frac{6}{5}}=  - \frac{1}{T},          
\]
where we use that $K\ll T^\frac{1}{3}$.

We got that the second derivative of $f$ is always negative. Therefore, $f$ is concave, and the $\beta$ value that minimize the expiration is at one of the edges of the domain $[0,T-\frac{K\tau}{4}+\tau]$, so it is sufficient to compare the value in $\beta=0$ and $\beta=T-(\frac{K}{4}-8)\tau$.

for $\beta = 0$ we have  
\begin{equation*}
\begin{split}        
\E _j^*[\Re] =
\frac{K^\frac{3}{5} T^\frac{4}{5}}{2} + \frac{K^\frac{3}{5}}{2T^\frac{1}{5}}.       
\end{split}
\end{equation*}

for $\beta = T-\frac{K\tau}{4}+8\tau$, 
\begin{equation*}
\begin{split}    
\E _j^*[\Re]   & =                       
\frac{(\frac{K\tau}{4}-8\tau)  (2T-\frac{K\tau}{4}+8\tau +1)}{2T} 
- \frac{(\frac{K\tau}{4}-8\tau) (\frac{K\tau}{4}-8\tau +1)}{2T}   
\\ & + \frac{K^\frac{3}{5} (\frac{K\tau}{4}-8\tau)(\frac{K\tau}{4}-8\tau+1)}{2T^\frac{6}{5}}              
\\ & =  
\frac{(\frac{K\tau}{4}-8\tau)  (T-\frac{K\tau}{4}+8\tau)}{T}          
+ \frac{K^\frac{3}{5} \overset{\geq 0}{\overbrace{(\frac{K\tau}{4}-\tau)^2}}}{2T^\frac{6}{5}}                                                    
+ \frac{K^\frac{8}{5} \tau}{8T^\frac{6}{5}}       
- \frac{4K^\frac{3}{5} \tau}{T^\frac{6}{5}}       
\\ & \geq                                          
\frac{K\tau}{4}
- \frac{K^2\tau^2}{16T}
+ \frac{4K\tau^2}{T}
-  8\tau
- \frac{64\tau^2}{T}
+ \frac{K^\frac{8}{5} \tau}{8T^\frac{6}{5}}       
- \frac{4K^\frac{3}{5} \tau}{T^\frac{6}{5}}       
\\ & =                                             
\frac{K^\frac{3}{5} T^\frac{4}{5}}{36}
- \frac{K^\frac{6}{5}T^\frac{3}{5} }{2304}
+ \frac{K^\frac{1}{5} T^\frac{3}{5}}{36}
-  \frac{3T^\frac{4}{5}}{4 K^\frac{2}{5}}
- \frac{2T^\frac{3}{5} }{5K^\frac{4}{5}}
+ \frac{K^\frac{6}{5} }{96T^\frac{2}{5}}       
- \frac{4K^\frac{1}{5}}{12T^\frac{2}{5}}       
\\ & \geq \frac{K^\frac{3}{5} T^\frac{4}{5}}{40},
\end{split}
\end{equation*}
where in the last inequality we use that $K \leq T^\frac{1}{3}$.

Finally we get: \[\E _j^*[\Re] \geq \frac{K^\frac{3}{5}T^\frac{4}{5}}{40}.\]
\end{proof}

\begin{document}

\maketitle

\begin{abstract}
  The abstract paragraph should be indented \nicefrac{1}{2}~inch (3~picas) on
  both the left- and right-hand margins. Use 10~point type, with a vertical
  spacing (leading) of 11~points.  The word \textbf{Abstract} must be centered,
  bold, and in point size 12. Two line spaces precede the abstract. The abstract
  must be limited to one paragraph.
\end{abstract}

\section{Submission of papers to NeurIPS 2024}

Please read the instructions below carefully and follow them faithfully.

\subsection{Style}

Papers to be submitted to NeurIPS 2024 must be prepared according to the
instructions presented here. Papers may only be up to {\bf nine} pages long,
including figures. Additional pages \emph{containing only acknowledgments and
references} are allowed. Papers that exceed the page limit will not be
reviewed, or in any other way considered for presentation at the conference.

The margins in 2024 are the same as those in previous years.

Authors are required to use the NeurIPS \LaTeX{} style files obtainable at the
NeurIPS website as indicated below. Please make sure you use the current files
and not previous versions. Tweaking the style files may be grounds for
rejection.

\subsection{Retrieval of style files}

The style files for NeurIPS and other conference information are available on
the website at
\begin{center}
  \url{http://www.neurips.cc/}
\end{center}
The file \verb+neurips_2024.pdf+ contains these instructions and illustrates the
various formatting requirements your NeurIPS paper must satisfy.

The only supported style file for NeurIPS 2024 is \verb+neurips_2024.sty+,
rewritten for \LaTeXe{}.  \textbf{Previous style files for \LaTeX{} 2.09,
  Microsoft Word, and RTF are no longer supported!}

The \LaTeX{} style file contains three optional arguments: \verb+final+, which
creates a camera-ready copy, \verb+preprint+, which creates a preprint for
submission to, e.g., arXiv, and \verb+nonatbib+, which will not load the
\verb+natbib+ package for you in case of package clash.

\paragraph{Preprint option}
If you wish to post a preprint of your work online, e.g., on arXiv, using the
NeurIPS style, please use the \verb+preprint+ option. This will create a
nonanonymized version of your work with the text ``Preprint. Work in progress.''
in the footer. This version may be distributed as you see fit, as long as you do not say which conference it was submitted to. Please \textbf{do
  not} use the \verb+final+ option, which should \textbf{only} be used for
papers accepted to NeurIPS.

At submission time, please omit the \verb+final+ and \verb+preprint+
options. This will anonymize your submission and add line numbers to aid
review. Please do \emph{not} refer to these line numbers in your paper as they
will be removed during generation of camera-ready copies.

The file \verb+neurips_2024.tex+ may be used as a ``shell'' for writing your
paper. All you have to do is replace the author, title, abstract, and text of
the paper with your own.

The formatting instructions contained in these style files are summarized in
Sections \ref{gen_inst}, \ref{headings}, and \ref{others} below.

\section{General formatting instructions}
\label{gen_inst}

The text must be confined within a rectangle 5.5~inches (33~picas) wide and
9~inches (54~picas) long. The left margin is 1.5~inch (9~picas).  Use 10~point
type with a vertical spacing (leading) of 11~points.  Times New Roman is the
preferred typeface throughout, and will be selected for you by default.
Paragraphs are separated by \nicefrac{1}{2}~line space (5.5 points), with no
indentation.

The paper title should be 17~point, initial caps/lower case, bold, centered
between two horizontal rules. The top rule should be 4~points thick and the
bottom rule should be 1~point thick. Allow \nicefrac{1}{4}~inch space above and
below the title to rules. All pages should start at 1~inch (6~picas) from the
top of the page.

For the final version, authors' names are set in boldface, and each name is
centered above the corresponding address. The lead author's name is to be listed
first (left-most), and the co-authors' names (if different address) are set to
follow. If there is only one co-author, list both author and co-author side by
side.

Please pay special attention to the instructions in Section \ref{others}
regarding figures, tables, acknowledgments, and references.

\section{Headings: first level}
\label{headings}

All headings should be lower case (except for first word and proper nouns),
flush left, and bold.

First-level headings should be in 12-point type.

\subsection{Headings: second level}

Second-level headings should be in 10-point type.

\subsubsection{Headings: third level}

Third-level headings should be in 10-point type.

\paragraph{Paragraphs}

There is also a \verb+\paragraph+ command available, which sets the heading in
bold, flush left, and inline with the text, with the heading followed by 1\,em
of space.

\section{Citations, figures, tables, references}
\label{others}

These instructions apply to everyone.

\subsection{Citations within the text}

The \verb+natbib+ package will be loaded for you by default.  Citations may be
author/year or numeric, as long as you maintain internal consistency.  As to the
format of the references themselves, any style is acceptable as long as it is
used consistently.

The documentation for \verb+natbib+ may be found at
\begin{center}
  \url{http://mirrors.ctan.org/macros/latex/contrib/natbib/natnotes.pdf}
\end{center}
Of note is the command \verb+\citet+, which produces citations appropriate for
use in inline text.  For example,
\begin{verbatim}
   \citet{hasselmo} investigated\dots
\end{verbatim}
produces
\begin{quote}
  Hasselmo, et al.\ (1995) investigated\dots
\end{quote}

If you wish to load the \verb+natbib+ package with options, you may add the
following before loading the \verb+neurips_2024+ package:
\begin{verbatim}
   \PassOptionsToPackage{options}{natbib}
\end{verbatim}

If \verb+natbib+ clashes with another package you load, you can add the optional
argument \verb+nonatbib+ when loading the style file:
\begin{verbatim}
   \usepackage[nonatbib]{neurips_2024}
\end{verbatim}

As submission is double blind, refer to your own published work in the third
person. That is, use ``In the previous work of Jones et al.\ [4],'' not ``In our
previous work [4].'' If you cite your other papers that are not widely available
(e.g., a journal paper under review), use anonymous author names in the
citation, e.g., an author of the form ``A.\ Anonymous'' and include a copy of the anonymized paper in the supplementary material.

\subsection{Footnotes}

Footnotes should be used sparingly.  If you do require a footnote, indicate
footnotes with a number\footnote{Sample of the first footnote.} in the
text. Place the footnotes at the bottom of the page on which they appear.
Precede the footnote with a horizontal rule of 2~inches (12~picas).

Note that footnotes are properly typeset \emph{after} punctuation
marks.\footnote{As in this example.}

\subsection{Figures}

\begin{figure}
  \centering
  \fbox{\rule[-.5cm]{0cm}{4cm} \rule[-.5cm]{4cm}{0cm}}
  \caption{Sample figure caption.}
\end{figure}

All artwork must be neat, clean, and legible. Lines should be dark enough for
purposes of reproduction. The figure number and caption always appear after the
figure. Place one line space before the figure caption and one line space after
the figure. The figure caption should be lower case (except for first word and
proper nouns); figures are numbered consecutively.

You may use color figures.  However, it is best for the figure captions and the
paper body to be legible if the paper is printed in either black/white or in
color.

\subsection{Tables}

All tables must be centered, neat, clean and legible.  The table number and
title always appear before the table.  See Table~\ref{sample-table}.

Place one line space before the table title, one line space after the
table title, and one line space after the table. The table title must
be lower case (except for first word and proper nouns); tables are
numbered consecutively.

Note that publication-quality tables \emph{do not contain vertical rules.} We
strongly suggest the use of the \verb+booktabs+ package, which allows for
typesetting high-quality, professional tables:
\begin{center}
  \url{https://www.ctan.org/pkg/booktabs}
\end{center}
This package was used to typeset Table~\ref{sample-table}.

\begin{table}
  \caption{Sample table title}
  \label{sample-table}
  \centering
  \begin{tabular}{lll}
    \toprule
    \multicolumn{2}{c}{Part}                   \\
    \cmidrule(r){1-2}
    Name     & Description     & Size ($\mu$m) \\
    \midrule
    Dendrite & Input terminal  & $\sim$100     \\
    Axon     & Output terminal & $\sim$10      \\
    Soma     & Cell body       & up to $10^6$  \\
    \bottomrule
  \end{tabular}
\end{table}

\subsection{Math}
Note that display math in bare TeX commands will not create correct line numbers for submission. Please use LaTeX (or AMSTeX) commands for unnumbered display math. (You really shouldn't be using \$\$ anyway; see \url{https://tex.stackexchange.com/questions/503/why-is-preferable-to} and \url{https://tex.stackexchange.com/questions/40492/what-are-the-differences-between-align-equation-and-displaymath} for more information.)

\subsection{Final instructions}

Do not change any aspects of the formatting parameters in the style files.  In
particular, do not modify the width or length of the rectangle the text should
fit into, and do not change font sizes (except perhaps in the
\textbf{References} section; see below). Please note that pages should be
numbered.

\section{Preparing PDF files}

Please prepare submission files with paper size ``US Letter,'' and not, for
example, ``A4.''

Fonts were the main cause of problems in the past years. Your PDF file must only
contain Type 1 or Embedded TrueType fonts. Here are a few instructions to
achieve this.

\begin{itemize}

\item You should directly generate PDF files using \verb+pdflatex+.

\item You can check which fonts a PDF files uses.  In Acrobat Reader, select the
  menu Files$>$Document Properties$>$Fonts and select Show All Fonts. You can
  also use the program \verb+pdffonts+ which comes with \verb+xpdf+ and is
  available out-of-the-box on most Linux machines.

\item \verb+xfig+ "patterned" shapes are implemented with bitmap fonts.  Use
  "solid" shapes instead.

\item The \verb+\bbold+ package almost always uses bitmap fonts.  You should use
  the equivalent AMS Fonts:
\begin{verbatim}
   \usepackage{amsfonts}
\end{verbatim}
followed by, e.g., \verb+\mathbb{R}+, \verb+\mathbb{N}+, or \verb+\mathbb{C}+
for $\mathbb{R}$, $\mathbb{N}$ or $\mathbb{C}$.  You can also use the following
workaround for reals, natural and complex:
\begin{verbatim}
   \newcommand{\RR}{I\!\!R} %real numbers
   \newcommand{\Nat}{I\!\!N} %natural numbers
   \newcommand{\CC}{I\!\!\!\!C} %complex numbers
\end{verbatim}
Note that \verb+amsfonts+ is automatically loaded by the \verb+amssymb+ package.

\end{itemize}

If your file contains type 3 fonts or non embedded TrueType fonts, we will ask
you to fix it.

\subsection{Margins in \LaTeX{}}

Most of the margin problems come from figures positioned by hand using
\verb+\special+ or other commands. We suggest using the command
\verb+\includegraphics+ from the \verb+graphicx+ package. Always specify the
figure width as a multiple of the line width as in the example below:
\begin{verbatim}
   \usepackage[pdftex]{graphicx} ...
   \includegraphics[width=0.8\linewidth]{myfile.pdf}
\end{verbatim}
See Section 4.4 in the graphics bundle documentation
(\url{http://mirrors.ctan.org/macros/latex/required/graphics/grfguide.pdf})

A number of width problems arise when \LaTeX{} cannot properly hyphenate a
line. Please give LaTeX hyphenation hints using the \verb+\-+ command when
necessary.

\begin{ack}
Use unnumbered first level headings for the acknowledgments. All acknowledgments
go at the end of the paper before the list of references. Moreover, you are required to declare
funding (financial activities supporting the submitted work) and competing interests (related financial activities outside the submitted work).
More information about this disclosure can be found at: \url{https://neurips.cc/Conferences/2024/PaperInformation/FundingDisclosure}.

Do {\bf not} include this section in the anonymized submission, only in the final paper. You can use the \texttt{ack} environment provided in the style file to automatically hide this section in the anonymized submission.
\end{ack}

\section*{References}

References follow the acknowledgments in the camera-ready paper. Use unnumbered first-level heading for
the references. Any choice of citation style is acceptable as long as you are
consistent. It is permissible to reduce the font size to \verb+small+ (9 point)
when listing the references.
Note that the Reference section does not count towards the page limit.
\medskip

{
\small

[1] Alexander, J.A.\ \& Mozer, M.C.\ (1995) Template-based algorithms for
connectionist rule extraction. In G.\ Tesauro, D.S.\ Touretzky and T.K.\ Leen
(eds.), {\it Advances in Neural Information Processing Systems 7},
pp.\ 609--616. Cambridge, MA: MIT Press.

[2] Bower, J.M.\ \& Beeman, D.\ (1995) {\it The Book of GENESIS: Exploring
  Realistic Neural Models with the GEneral NEural SImulation System.}  New York:
TELOS/Springer--Verlag.

[3] Hasselmo, M.E., Schnell, E.\ \& Barkai, E.\ (1995) Dynamics of learning and
recall at excitatory recurrent synapses and cholinergic modulation in rat
hippocampal region CA3. {\it Journal of Neuroscience} {\bf 15}(7):5249-5262.
}


\appendix

\section{Appendix / supplemental material}

Optionally include supplemental material (complete proofs, additional experiments and plots) in appendix.
All such materials \textbf{SHOULD be included in the main submission.}


\newpage
\section*{NeurIPS Paper Checklist}

The checklist is designed to encourage best practices for responsible machine learning research, addressing issues of reproducibility, transparency, research ethics, and societal impact. Do not remove the checklist: {\bf The papers not including the checklist will be desk rejected.} The checklist should follow the references and follow the (optional) supplemental material.  The checklist does NOT count towards the page
limit. 

Please read the checklist guidelines carefully for information on how to answer these questions. For each question in the checklist:
\begin{itemize}
    \item You should answer \answerYes{}, \answerNo{}, or \answerNA{}.
    \item \answerNA{} means either that the question is Not Applicable for that particular paper or the relevant information is Not Available.
    \item Please provide a short (1–2 sentence) justification right after your answer (even for NA). 
\end{itemize}

{\bf The checklist answers are an integral part of your paper submission.} They are visible to the reviewers, area chairs, senior area chairs, and ethics reviewers. You will be asked to also include it (after eventual revisions) with the final version of your paper, and its final version will be published with the paper.

The reviewers of your paper will be asked to use the checklist as one of the factors in their evaluation. While "\answerYes{}" is generally preferable to "\answerNo{}", it is perfectly acceptable to answer "\answerNo{}" provided a proper justification is given (e.g., "error bars are not reported because it would be too computationally expensive" or "we were unable to find the license for the dataset we used"). In general, answering "\answerNo{}" or "\answerNA{}" is not grounds for rejection. While the questions are phrased in a binary way, we acknowledge that the true answer is often more nuanced, so please just use your best judgment and write a justification to elaborate. All supporting evidence can appear either in the main paper or the supplemental material, provided in appendix. If you answer \answerYes{} to a question, in the justification please point to the section(s) where related material for the question can be found.

IMPORTANT, please:
\begin{itemize}
    \item {\bf Delete this instruction block, but keep the section heading ``NeurIPS paper checklist"},
    \item  {\bf Keep the checklist subsection headings, questions/answers and guidelines below.}
    \item {\bf Do not modify the questions and only use the provided macros for your answers}.
\end{itemize}


\begin{enumerate}

\item {\bf Claims}
    \item[] Question: Do the main claims made in the abstract and introduction accurately reflect the paper's contributions and scope?
    \item[] Answer: \answerTODO{} 
    \item[] Justification: \justificationTODO{}
    \item[] Guidelines:
    \begin{itemize}
        \item The answer NA means that the abstract and introduction do not include the claims made in the paper.
        \item The abstract and/or introduction should clearly state the claims made, including the contributions made in the paper and important assumptions and limitations. A No or NA answer to this question will not be perceived well by the reviewers. 
        \item The claims made should match theoretical and experimental results, and reflect how much the results can be expected to generalize to other settings. 
        \item It is fine to include aspirational goals as motivation as long as it is clear that these goals are not attained by the paper. 
    \end{itemize}

\item {\bf Limitations}
    \item[] Question: Does the paper discuss the limitations of the work performed by the authors?
    \item[] Answer: \answerTODO{} 
    \item[] Justification: \justificationTODO{}
    \item[] Guidelines:
    \begin{itemize}
        \item The answer NA means that the paper has no limitation while the answer No means that the paper has limitations, but those are not discussed in the paper. 
        \item The authors are encouraged to create a separate "Limitations" section in their paper.
        \item The paper should point out any strong assumptions and how robust the results are to violations of these assumptions (e.g., independence assumptions, noiseless settings, model well-specification, asymptotic approximations only holding locally). The authors should reflect on how these assumptions might be violated in practice and what the implications would be.
        \item The authors should reflect on the scope of the claims made, e.g., if the approach was only tested on a few datasets or with a few runs. In general, empirical results often depend on implicit assumptions, which should be articulated.
        \item The authors should reflect on the factors that influence the performance of the approach. For example, a facial recognition algorithm may perform poorly when image resolution is low or images are taken in low lighting. Or a speech-to-text system might not be used reliably to provide closed captions for online lectures because it fails to handle technical jargon.
        \item The authors should discuss the computational efficiency of the proposed algorithms and how they scale with dataset size.
        \item If applicable, the authors should discuss possible limitations of their approach to address problems of privacy and fairness.
        \item While the authors might fear that complete honesty about limitations might be used by reviewers as grounds for rejection, a worse outcome might be that reviewers discover limitations that aren't acknowledged in the paper. The authors should use their best judgment and recognize that individual actions in favor of transparency play an important role in developing norms that preserve the integrity of the community. Reviewers will be specifically instructed to not penalize honesty concerning limitations.
    \end{itemize}

\item {\bf Theory Assumptions and Proofs}
    \item[] Question: For each theoretical result, does the paper provide the full set of assumptions and a complete (and correct) proof?
    \item[] Answer: \answerTODO{} 
    \item[] Justification: \justificationTODO{}
    \item[] Guidelines:
    \begin{itemize}
        \item The answer NA means that the paper does not include theoretical results. 
        \item All the theorems, formulas, and proofs in the paper should be numbered and cross-referenced.
        \item All assumptions should be clearly stated or referenced in the statement of any theorems.
        \item The proofs can either appear in the main paper or the supplemental material, but if they appear in the supplemental material, the authors are encouraged to provide a short proof sketch to provide intuition. 
        \item Inversely, any informal proof provided in the core of the paper should be complemented by formal proofs provided in appendix or supplemental material.
        \item Theorems and Lemmas that the proof relies upon should be properly referenced. 
    \end{itemize}

    \item {\bf Experimental Result Reproducibility}
    \item[] Question: Does the paper fully disclose all the information needed to reproduce the main experimental results of the paper to the extent that it affects the main claims and/or conclusions of the paper (regardless of whether the code and data are provided or not)?
    \item[] Answer: \answerTODO{} 
    \item[] Justification: \justificationTODO{}
    \item[] Guidelines:
    \begin{itemize}
        \item The answer NA means that the paper does not include experiments.
        \item If the paper includes experiments, a No answer to this question will not be perceived well by the reviewers: Making the paper reproducible is important, regardless of whether the code and data are provided or not.
        \item If the contribution is a dataset and/or model, the authors should describe the steps taken to make their results reproducible or verifiable. 
        \item Depending on the contribution, reproducibility can be accomplished in various ways. For example, if the contribution is a novel architecture, describing the architecture fully might suffice, or if the contribution is a specific model and empirical evaluation, it may be necessary to either make it possible for others to replicate the model with the same dataset, or provide access to the model. In general. releasing code and data is often one good way to accomplish this, but reproducibility can also be provided via detailed instructions for how to replicate the results, access to a hosted model (e.g., in the case of a large language model), releasing of a model checkpoint, or other means that are appropriate to the research performed.
        \item While NeurIPS does not require releasing code, the conference does require all submissions to provide some reasonable avenue for reproducibility, which may depend on the nature of the contribution. For example
        \begin{enumerate}
            \item If the contribution is primarily a new algorithm, the paper should make it clear how to reproduce that algorithm.
            \item If the contribution is primarily a new model architecture, the paper should describe the architecture clearly and fully.
            \item If the contribution is a new model (e.g., a large language model), then there should either be a way to access this model for reproducing the results or a way to reproduce the model (e.g., with an open-source dataset or instructions for how to construct the dataset).
            \item We recognize that reproducibility may be tricky in some cases, in which case authors are welcome to describe the particular way they provide for reproducibility. In the case of closed-source models, it may be that access to the model is limited in some way (e.g., to registered users), but it should be possible for other researchers to have some path to reproducing or verifying the results.
        \end{enumerate}
    \end{itemize}

\item {\bf Open access to data and code}
    \item[] Question: Does the paper provide open access to the data and code, with sufficient instructions to faithfully reproduce the main experimental results, as described in supplemental material?
    \item[] Answer: \answerTODO{} 
    \item[] Justification: \justificationTODO{}
    \item[] Guidelines:
    \begin{itemize}
        \item The answer NA means that paper does not include experiments requiring code.
        \item Please see the NeurIPS code and data submission guidelines (\url{https://nips.cc/public/guides/CodeSubmissionPolicy}) for more details.
        \item While we encourage the release of code and data, we understand that this might not be possible, so “No” is an acceptable answer. Papers cannot be rejected simply for not including code, unless this is central to the contribution (e.g., for a new open-source benchmark).
        \item The instructions should contain the exact command and environment needed to run to reproduce the results. See the NeurIPS code and data submission guidelines (\url{https://nips.cc/public/guides/CodeSubmissionPolicy}) for more details.
        \item The authors should provide instructions on data access and preparation, including how to access the raw data, preprocessed data, intermediate data, and generated data, etc.
        \item The authors should provide scripts to reproduce all experimental results for the new proposed method and baselines. If only a subset of experiments are reproducible, they should state which ones are omitted from the script and why.
        \item At submission time, to preserve anonymity, the authors should release anonymized versions (if applicable).
        \item Providing as much information as possible in supplemental material (appended to the paper) is recommended, but including URLs to data and code is permitted.
    \end{itemize}

\item {\bf Experimental Setting/Details}
    \item[] Question: Does the paper specify all the training and test details (e.g., data splits, hyperparameters, how they were chosen, type of optimizer, etc.) necessary to understand the results?
    \item[] Answer: \answerTODO{} 
    \item[] Justification: \justificationTODO{}
    \item[] Guidelines:
    \begin{itemize}
        \item The answer NA means that the paper does not include experiments.
        \item The experimental setting should be presented in the core of the paper to a level of detail that is necessary to appreciate the results and make sense of them.
        \item The full details can be provided either with the code, in appendix, or as supplemental material.
    \end{itemize}

\item {\bf Experiment Statistical Significance}
    \item[] Question: Does the paper report error bars suitably and correctly defined or other appropriate information about the statistical significance of the experiments?
    \item[] Answer: \answerTODO{} 
    \item[] Justification: \justificationTODO{}
    \item[] Guidelines:
    \begin{itemize}
        \item The answer NA means that the paper does not include experiments.
        \item The authors should answer "Yes" if the results are accompanied by error bars, confidence intervals, or statistical significance tests, at least for the experiments that support the main claims of the paper.
        \item The factors of variability that the error bars are capturing should be clearly stated (for example, train/test split, initialization, random drawing of some parameter, or overall run with given experimental conditions).
        \item The method for calculating the error bars should be explained (closed form formula, call to a library function, bootstrap, etc.)
        \item The assumptions made should be given (e.g., Normally distributed errors).
        \item It should be clear whether the error bar is the standard deviation or the standard error of the mean.
        \item It is OK to report 1-sigma error bars, but one should state it. The authors should preferably report a 2-sigma error bar than state that they have a 96\% CI, if the hypothesis of Normality of errors is not verified.
        \item For asymmetric distributions, the authors should be careful not to show in tables or figures symmetric error bars that would yield results that are out of range (e.g. negative error rates).
        \item If error bars are reported in tables or plots, The authors should explain in the text how they were calculated and reference the corresponding figures or tables in the text.
    \end{itemize}

\item {\bf Experiments Compute Resources}
    \item[] Question: For each experiment, does the paper provide sufficient information on the computer resources (type of compute workers, memory, time of execution) needed to reproduce the experiments?
    \item[] Answer: \answerTODO{} 
    \item[] Justification: \justificationTODO{}
    \item[] Guidelines:
    \begin{itemize}
        \item The answer NA means that the paper does not include experiments.
        \item The paper should indicate the type of compute workers CPU or GPU, internal cluster, or cloud provider, including relevant memory and storage.
        \item The paper should provide the amount of compute required for each of the individual experimental runs as well as estimate the total compute. 
        \item The paper should disclose whether the full research project required more compute than the experiments reported in the paper (e.g., preliminary or failed experiments that didn't make it into the paper). 
    \end{itemize}
    
\item {\bf Code Of Ethics}
    \item[] Question: Does the research conducted in the paper conform, in every respect, with the NeurIPS Code of Ethics \url{https://neurips.cc/public/EthicsGuidelines}?
    \item[] Answer: \answerTODO{} 
    \item[] Justification: \justificationTODO{}
    \item[] Guidelines:
    \begin{itemize}
        \item The answer NA means that the authors have not reviewed the NeurIPS Code of Ethics.
        \item If the authors answer No, they should explain the special circumstances that require a deviation from the Code of Ethics.
        \item The authors should make sure to preserve anonymity (e.g., if there is a special consideration due to laws or regulations in their jurisdiction).
    \end{itemize}

\item {\bf Broader Impacts}
    \item[] Question: Does the paper discuss both potential positive societal impacts and negative societal impacts of the work performed?
    \item[] Answer: \answerTODO{} 
    \item[] Justification: \justificationTODO{}
    \item[] Guidelines:
    \begin{itemize}
        \item The answer NA means that there is no societal impact of the work performed.
        \item If the authors answer NA or No, they should explain why their work has no societal impact or why the paper does not address societal impact.
        \item Examples of negative societal impacts include potential malicious or unintended uses (e.g., disinformation, generating fake profiles, surveillance), fairness considerations (e.g., deployment of technologies that could make decisions that unfairly impact specific groups), privacy considerations, and security considerations.
        \item The conference expects that many papers will be foundational research and not tied to particular applications, let alone deployments. However, if there is a direct path to any negative applications, the authors should point it out. For example, it is legitimate to point out that an improvement in the quality of generative models could be used to generate deepfakes for disinformation. On the other hand, it is not needed to point out that a generic algorithm for optimizing neural networks could enable people to train models that generate Deepfakes faster.
        \item The authors should consider possible harms that could arise when the technology is being used as intended and functioning correctly, harms that could arise when the technology is being used as intended but gives incorrect results, and harms following from (intentional or unintentional) misuse of the technology.
        \item If there are negative societal impacts, the authors could also discuss possible mitigation strategies (e.g., gated release of models, providing defenses in addition to attacks, mechanisms for monitoring misuse, mechanisms to monitor how a system learns from feedback over time, improving the efficiency and accessibility of ML).
    \end{itemize}
    
\item {\bf Safeguards}
    \item[] Question: Does the paper describe safeguards that have been put in place for responsible release of data or models that have a high risk for misuse (e.g., pretrained language models, image generators, or scraped datasets)?
    \item[] Answer: \answerTODO{} 
    \item[] Justification: \justificationTODO{}
    \item[] Guidelines:
    \begin{itemize}
        \item The answer NA means that the paper poses no such risks.
        \item Released models that have a high risk for misuse or dual-use should be released with necessary safeguards to allow for controlled use of the model, for example by requiring that users adhere to usage guidelines or restrictions to access the model or implementing safety filters. 
        \item Datasets that have been scraped from the Internet could pose safety risks. The authors should describe how they avoided releasing unsafe images.
        \item We recognize that providing effective safeguards is challenging, and many papers do not require this, but we encourage authors to take this into account and make a best faith effort.
    \end{itemize}

\item {\bf Licenses for existing assets}
    \item[] Question: Are the creators or original owners of assets (e.g., code, data, models), used in the paper, properly credited and are the license and terms of use explicitly mentioned and properly respected?
    \item[] Answer: \answerTODO{} 
    \item[] Justification: \justificationTODO{}
    \item[] Guidelines:
    \begin{itemize}
        \item The answer NA means that the paper does not use existing assets.
        \item The authors should cite the original paper that produced the code package or dataset.
        \item The authors should state which version of the asset is used and, if possible, include a URL.
        \item The name of the license (e.g., CC-BY 4.0) should be included for each asset.
        \item For scraped data from a particular source (e.g., website), the copyright and terms of service of that source should be provided.
        \item If assets are released, the license, copyright information, and terms of use in the package should be provided. For popular datasets, \url{paperswithcode.com/datasets} has curated licenses for some datasets. Their licensing guide can help determine the license of a dataset.
        \item For existing datasets that are re-packaged, both the original license and the license of the derived asset (if it has changed) should be provided.
        \item If this information is not available online, the authors are encouraged to reach out to the asset's creators.
    \end{itemize}

\item {\bf New Assets}
    \item[] Question: Are new assets introduced in the paper well documented and is the documentation provided alongside the assets?
    \item[] Answer: \answerTODO{} 
    \item[] Justification: \justificationTODO{}
    \item[] Guidelines:
    \begin{itemize}
        \item The answer NA means that the paper does not release new assets.
        \item Researchers should communicate the details of the dataset/code/model as part of their submissions via structured templates. This includes details about training, license, limitations, etc. 
        \item The paper should discuss whether and how consent was obtained from people whose asset is used.
        \item At submission time, remember to anonymize your assets (if applicable). You can either create an anonymized URL or include an anonymized zip file.
    \end{itemize}

\item {\bf Crowdsourcing and Research with Human Subjects}
    \item[] Question: For crowdsourcing experiments and research with human subjects, does the paper include the full text of instructions given to participants and screenshots, if applicable, as well as details about compensation (if any)? 
    \item[] Answer: \answerTODO{} 
    \item[] Justification: \justificationTODO{}
    \item[] Guidelines:
    \begin{itemize}
        \item The answer NA means that the paper does not involve crowdsourcing nor research with human subjects.
        \item Including this information in the supplemental material is fine, but if the main contribution of the paper involves human subjects, then as much detail as possible should be included in the main paper. 
        \item According to the NeurIPS Code of Ethics, workers involved in data collection, curation, or other labor should be paid at least the minimum wage in the country of the data collector. 
    \end{itemize}

\item {\bf Institutional Review Board (IRB) Approvals or Equivalent for Research with Human Subjects}
    \item[] Question: Does the paper describe potential risks incurred by study participants, whether such risks were disclosed to the subjects, and whether Institutional Review Board (IRB) approvals (or an equivalent approval/review based on the requirements of your country or institution) were obtained?
    \item[] Answer: \answerTODO{} 
    \item[] Justification: \justificationTODO{}
    \item[] Guidelines:
    \begin{itemize}
        \item The answer NA means that the paper does not involve crowdsourcing nor research with human subjects.
        \item Depending on the country in which research is conducted, IRB approval (or equivalent) may be required for any human subjects research. If you obtained IRB approval, you should clearly state this in the paper. 
        \item We recognize that the procedures for this may vary significantly between institutions and locations, and we expect authors to adhere to the NeurIPS Code of Ethics and the guidelines for their institution. 
        \item For initial submissions, do not include any information that would break anonymity (if applicable), such as the institution conducting the review.
    \end{itemize}

\end{enumerate}

\end{document}